\documentclass[sn-mathphys,Numbered]{sn-jnl}

\usepackage{upquote}
\usepackage{amsmath,amssymb,amsfonts}
\usepackage{graphicx}
\usepackage{textcomp}
\usepackage[utf8]{inputenc}
\usepackage{xcolor}
\usepackage{adjustbox} 
\usepackage{makecell} 
\usepackage[ngerman,main=english]{babel}
\usepackage{algorithm}
\usepackage{algorithmic}
\usepackage{multirow}
\usepackage{caption}
\usepackage{subfloat}
\usepackage{pgfplots}
\usepackage{tabularx}
\usepackage{array}
\usepackage{booktabs}
\usepackage{graphicx}%
\usepackage{adjustbox} 
\usepackage{subcaption}
\usepackage{multirow}%
\usepackage{amsmath,amssymb,amsfonts}%
\usepackage{amsthm}%
\usepackage{mathrsfs}%
\usepackage[title]{appendix}%
\usepackage{xcolor}%
\usepackage{textcomp}%
\usepackage{manyfoot}%
\usepackage{booktabs}%
\usepackage{listings}%

\begin{document}

\title[Article Title]{Securing Recommender System via Cooperative Training}


\author[1,2]{\fnm{Qingyang} \sur{Wang}}\email{greensun@mail.ustc.edu.cn}
\equalcont{These authors contributed equally to this work.}

\author[1,2]{\fnm{Chenwang} \sur{Wu}}\email{wcw1996@mail.ustc.edu.cn}
\equalcont{These authors contributed equally to this work.}

\author*[1,2]{\fnm{Defu} \sur{Lian}}\email{liandefu@ustc.edu.cn}

\author[1,2]{\fnm{Enhong} \sur{Chen}}\email{cheneh@ustc.edu.cn}

\affil[1]{\orgdiv{School of Data Science}, \orgname{University of Science and Technology of China}, \orgaddress{\street{96 Jinzhai Road}, \city{Hefei}, \state{Anhui}, \country{China}}}
\affil[2]{\orgdiv{State Key Laboratory of Cognitive Intelligence}, \orgaddress{\street{96 Jinzhai Road}, \city{Hefei}, \state{Anhui}, \country{China}}}


\abstract{Recommender systems are often susceptible to well-crafted fake profiles, leading to biased recommendations. Among existing defense methods, data-processing-based methods inevitably exclude normal samples, while model-based methods struggle to enjoy both generalization and robustness. To this end, we suggest integrating data processing and the robust model to propose a general framework, Triple Cooperative Defense (TCD), which employs three cooperative models that mutually enhance data and thereby improve recommendation robustness. Furthermore, Considering that existing attacks struggle to balance bi-level optimization and efficiency, we revisit poisoning attacks in recommender systems and introduce an efficient attack strategy, Co-training Attack (Co-Attack), which cooperatively optimizes the attack optimization and model training, considering the bi-level setting while maintaining attack efficiency. Moreover, we reveal a potential reason for the insufficient threat of existing attacks is their default assumption of optimizing attacks in undefended scenarios. This overly optimistic setting limits the potential of attacks. Consequently, we put forth a Game-based Co-training Attack (GCoAttack), which frames the proposed CoAttack and TCD as a game-theoretic process, thoroughly exploring CoAttack's attack potential in the cooperative training of attack and defense. Extensive experiments on three real datasets demonstrate TCD's superiority in enhancing model robustness. Additionally, we verify that the two proposed attack strategies significantly outperform existing attacks, with game-based GCoAttack posing a greater poisoning threat than CoAttack.}

\keywords{Recommender Systems, Model Robustness, Poisoning Attacks.}



\maketitle

\section{Introduction}\label{sec1}

In recent years, we have witnessed explosive growth in the amount of available information due to the rapid development of Internet technology. To cope with this massive data, ``recommender system''\cite{bobadilla2013recommender} has become a popular tool for quickly and effectively obtaining valuable information, gaining extensive attention in academia and industry. They mine the content that the user is interested in from a large amount of data by using information such as user behavior and item characteristics and presenting it to the user in a list\cite{himeur2022blockchain}. Benefiting their effectiveness and commercial background, they are widely used across various industries such as geographic recommendation \cite{lian2020geography}, e-commerce \cite{chevalier2006effect}, and audio-visual entertainment \cite{chevalier2006effect}.

However, recommender systems also face severe security challenges while providing convenience for our lives. This is because collaborative filtering, which is adopted in many recommender systems and recommends items based on user profile information, is vulnerable to fake user profiles. Several studies \cite{wu2021triple, li2016data, lin2020attacking} have demonstrated that recommender systems, especially those in the sales and rating domains, are systematically affected by user ratings within the system. This interference can impact users' purchase behavior and the system's recommendation outcomes \cite{chevalier2006effect}. Furthermore, it is worth emphasizing even if attackers do not know the algorithm or implementation details used by the recommendation system, only using small-scale misleading data can also have obvious interference effects on the normal recommendation behavior of the system (e.g., in 2002,  after receiving a complaint, Amazon found that when a website recommends a Christian classic, another irrelevant book will be recommended simultaneously, which is caused by malicious users using deceptive means \cite{liu2014new}).

Existing defense strategies against poisoning attacks in recommendations are mainly divided into (1) data-processing-based defense, which involves studying the features of poisoning profiles, removing fake profiles, and refining datasets before recommender system training. However, these methods may delete normal data to achieve high recall, leading to biased recommendations. (2) Model-based defense aims to enhance the recommendation algorithm's robustness even if fake data exists. Adversarial training \cite{madry2017towards} is an effective model-based defense method that maximizes recommendation error while minimizing the model's empirical risk by adding adversarial perturbations to the model parameters, eventually building robust models in adversarial games. While adversarial training can significantly improve the system's robustness, controlling the strength of adversarial noise is challenging, leading to reduced recommendation generalization. Moreover, recent research \cite{wu2021fight} suggests that adversarial training with perturbations added to model parameters may not resist poisoning attacks effectively. Thus, it is crucial to integrate both methods while leveraging their strengths and avoiding weaknesses.

Considering the shortcomings mentioned above, we propose a novel defense method, Triple Cooperative Defense (TCD), to enhance the robustness of recommender systems by integrating data processing and model robustness boosting. In TCD, three recommendation models are used for cooperative training. Specifically, in each round of training, the high-confidence prediction ratings of any two models are used as auxiliary training data for the remaining model. The three models cooperatively improve recommendation robustness. Our strategy is based on the following considerations. In the recommender system, extremely sparse user-item interactions (indicating less training data) are difficult to support good model training, leading to models easily misled by malicious profiles. Besides, recent work \cite{wu2021fight} also emphasizes that the model's robustness requires more real data. Therefore, we reasonably use cheap pseudo-labels (predicted ratings), whose reliability is guaranteed by the rating's confidence. To obtain high confidence ratings, we based on prior knowledge that ratings predicted to be consistent by most models are more credible. More model votes can lead to more reliable pseudo-labels, but at the cost of increased computation. Therefore, we compromise and suggest training with three models and any two models' consistent prediction ratings as auxiliary training data for the third model. Model robustness is improved in data augmentation and co-training of the three models. Notably, we do not cull the data or modify the individual model structure, which can overcome the shortcomings of the existing defense methods discussed above. 

Additionally, we investigate poisoning attacks in recommender systems. Existing attacks can be categorized into heuristic-based and optimization-based attacks \cite{nguyen2023poisoning}. Heuristic-based attacks, such as Average attack \cite{lam2004shilling} and Bandwagon attack \cite{burke2006classification}, design fake users based on the feature that similar users have similar interests in recommendations. However, they do not formally analyze specific recommendation models and cannot cover all recommendation patterns, resulting in insufficient attack performance. Consequently, model-based attacks have emerged. These methods target specific recommender systems, design optimization objective functions for fake profile generation, and use the projected gradient ascent to optimize the attack model, including visually-aware recommender systems \cite{cohen2021black}, sequence-based recommender systems \cite{yue2021black}, federated recommender systems \cite{zhang2022pipattack}, graph-based recommender systems \cite{fang2018poisoning}. Nevertheless, most of them assume the invariance of default parameters and do not consider the bi-level setting of poisoning attacks, that is, the optimization of poisoned data is accompanied by changes in model parameters. As a result, the optimized attack models they obtain are not optimal, reducing the destructive power of the attacks.

In response to these issues, Huang et al. \cite{huang2021data} proposed a poisoning attack that integrates model training and attack optimization. It constructs a poisoning model to simulate a target recommender system and updates the poisoning model based on the attack objectives, allowing it to approximate the state of the victim recommendation model after training. Although it considers bi-level optimization of poisoning and demonstrates satisfactory performance, it also has some shortcomings: (1) Low efficiency. It generates only one fake profile at a time, with each user requiring a complete recommendation training process. This means that poisoning $n'$ users necessitates training the model $n'$ times. This extremely time-consuming generation strategy limits its practicability on large-scale datasets. (2) Non-global optimal. It adopts a greedy strategy to optimize poisoning profiles one at a time, which means that when optimizing the $n$-th user, the first $n-1$ users remain fixed. Such constraint limits the search space and makes the algorithm prone to being trapped in local optima.

Inspired by them and considering the above mentioned deficiencies, we propose an efficient attack, Co-training Attack (CoAttack). We also combine attack optimization and model training for cooperative training, but the difference is that our optimization is based on all candidate poisoning data. This design can well solve their limitations: (1) the optimization based on all poisoned data only needs one complete recommendation training, which is efficient; (2) such optimization strategy can search in the complete feasible space, which is more helpful in finding the global optimal solution. Specifically, in CoAttack, we first initialize poisoning profiles and inject them into the target model. Then we pre-train the recommendation model based on the standard recommendation loss to ensure its ability to learn user preferences accurately. Third, we combine attack loss and recommendation loss for training, striving to approximate the attack model to the target model after poisoning. Finally, we choose the items with high predicted scores in the optimized model as filler items.

In addition, it is widely acknowledged that the dynamics of attack and defense in recommender systems resemble an ongoing arms race, where a previously effective attack strategy eventually becomes ineffective due to evolving defense mechanisms. One possible reason for the failure of attacks lies in the assumption that the target model remains defenseless and unoptimized, which is often overly optimistic.
To this end, we frame the interaction between attack and defense as a strategic game and train them in a coordinated manner. In this paper, we combine our proposed Triple Cooperative Defense (TCD) and CoAttack methodologies to introduce a novel approach called Game-based Co-training Attack (GCoAttack). In each round of attack optimization, GCoAttack first enhances recommendation robustness through TCD, then fine-tunes poisoning profiles using the robust recommendation model, and ultimately seeks to maximize attack performance within the context of this zero-sum game.

Except for the contributions in the preliminary work \cite{wang2022towards} on triple cooperative defense, we further deliver the following contributions:
\begin{itemize}
    \item We develop an effective poisoning attack method, CoAttack, which could efficiently generate malicious poisoning profiles by combining attack optimization and model training for cooperative training.
    \item We reveal the importance of games in the robust recommendation and propose GCoAttack. It further explores the attack potential of CoAttack through the cooperative training of CoAttack and TCD.
    \item Extensive experiments on three different datasets demonstrate the effectiveness of CoAttack (cooperative training between attack optimization and model training) and GCoAttack (cooperative training between attack and defense) over the state-of-the-art methods.
\end{itemize}

The rest of the paper is organized as follows: Section \ref{sec2} introduces related work. Section \ref{sec3} describes the threat model for recommendation poisoning. Sections \ref{sec4} and \ref{sec5} respectively present the proposed defense strategy TCD and two cooperative training attacks, CoAttack and GCoAttack. Section \ref{sec6} provides a comprehensive experimental comparison and analysis. The final section\ref{sec7} summarizes our work and looks forward to the future.

\section{Related Work}\label{sec2}

\subsection{Poisoning Attacks in Recommender Systems}
Many issues about security and privacy have been studied in recommender systems. These researches have revealed vulnerabilities in recommender systems \cite{du2018enhancing, si2020shilling}, prompting the development of a dedicated toolkit for assessing their robustness \cite{ovaisi2022rgrecsys}.
Earlier attacks injected malicious profiles manually generated with little knowledge about the recommender system,
so it could not achieve satisfactory attack performance, e.g., random
attack\cite{lam2004shilling} and average attack \cite{lam2004shilling}. 
To overcome this limitation, recent attacks have been optimized for specific classes of RSs, such as matrix-factorization-based\cite{li2016data}, neighborhood-based\cite{chen2021data}, graph-based\cite{fang2018poisoning}, deep-learning-based\cite{huang2021data}, and visual-awareness-based RSs\cite{cohen2021black}.
The training of these model-based recommendation algorithms usually
used backpropagation \cite{guo2017deepfm, he2017neural}, so perturbations were added along
the gradient direction to perform the attack \cite{fang2020influence, fang2018poisoning,li2016data,tang2020revisiting}. For example,  Nguyen et al.\cite{nguyen2023poisoning} studied poisoning attacks on GNN-based recommenders and proposed a solution involving surrogating a recommendation model and generating fake users and user-item interactions while preserving correlations. Inspired
by the GAN’s application \cite{jin2020sampling} in the recommendation, some works\cite{christakopoulou2019adversarial, lin2020attacking} used GAN to generate real-like fake ratings to bypass the detection. To address traditional GAN's inability to evaluate attacking damage, Wu et al. \cite{wu2021triple} drew inspiration from TripleGAN \cite{li2017triple} and introduced an attack module that conducts triple adversarial learning to generate malicious users. Advancements in deep learning have led Huang et al. \cite{huang2021data} to propose a poison attack on DL-based RSs. The attack involves injecting fake users with carefully crafted ratings into the targeted model and can be defined as an optimization problem that can be efficiently resolved by incorporating heuristic rules. The attack model remains practical even when attackers have limited access to only a small portion of user-item interactions. With
the development of optimization algorithms, many works focused on attacking specific types of recommender systems and turned attacks into optimization problems of deciding appropriate rating scores for users \cite{lam2004shilling,li2016data,yang2017fake,fang2018poisoning, oh2022robustness}. Moreover, some works\cite{fan2021attacking,song2020poisonrec}treated the items’ ratings as actions and used reinforcement learning to generate real-like fake ratings. Such optimization-based methods have strong attack performance, so defense is needed to mitigate the harm of attack. 

\subsection{Defense against Poisoning Attacks}
According to the defense objective, a defense can be (i) reactive attack detection\cite{deldjoo2021survey} or (ii) proactive robust model construction, which will be listed below.

\textbf{Reactive Attack detection.} 
Many works have been developed to detect shilling attacks\cite{si2020shilling,yang2016re,ge2022survey,zhang2015catch}, and such methods can be roughly classified into supervised classification methods and unsupervised clustering methods. The majority of work on supervised classification methods begins
with feature engineering and then turns to the development of algorithms. For example, Yang et al.\cite{yang2016re} utilized three carefully crafted features derived from user profiles for identifying attack profiles, which are the maximum, minimum, and average ratings of filler size. Besides, many researchers used KNN, C4.5, and SVM \cite{burke2006classification}to supervise the statistical attributes to detect attacks. In most practical recommendation systems, due to the small number of labeled users and the lack of prior knowledge, unsupervised learning \cite{zhang2018ud,zhang2014detection}and
semi-supervised learning \cite{cao2013shilling} were used to detect attacks. Unsupervised clustering methods usually aim to group individuals into
groups and then eliminate suspicious ones. For example, adversarial sample detection techniques utilize machine learning methods, such as principal component analysis (PCA)\cite{cheng2009effective}, to extract content features to identify adversarial samples. However, to pursue high recall, these methods inevitably delete normal data, which leads to biased recommendations. Conversely, for our proposed TCD to enrich high-confidence data rather than remove outliers, it can avoid cleaning normal data and train a more accurate and robust model.

\textbf{Proactive Robust recommendation.} Athalye et al.\cite{athalye2018obfuscated} proposed defenses based on gradient masking to produce models containing smoother gradients that hinder
optimization-based attack algorithms from finding the wrong directions in space\cite{machado2021adversarial}. More recently,
many works\cite{du2018enhancing,he2018adversarial,li2020adversarial,park2019adversarial,tang2019adversarial,yue2022defending} have focused on adversarial
training. Assuming that each instance may be the target of attacks
\cite{machado2021adversarial}, adversarial training adds perturbations to the inputs or model
parameters that force the model to learn fragile perturbations. Although adversarial training can significantly improve the robustness of recommender systems, it is difficult to control the strength of adversarial data, reducing the generalization of the recommendation. Instead, our proposed TCD does not require the addition of sensitive noise and is trained cooperatively to facilitate generalization, as we will demonstrate in Section \ref{sec4}.
In addition, many researchers have also improved the robustness of models by altering the way matrix factorization is performed. Hidano and Kiyomoto \cite{hidano2020recommender} propose a defensive strategy that utilizes trim learning to enhance the resilience of matrix factorization against data poisoning. This approach leverages the statistical difference between normal and fake users to protect against malicious manipulation of the data. Zhang et al. \cite{zhang2017robust} proposed a robust collaborative filtering method incorporating non-negative matrix factorization (NMF) with R1-norm. While in \cite{yu2017novel}, the authors have designed a robust matrix factorization model based on kernel mapping and kernel distance.

\section{Threat Model}\label{sec3}

In this section, we formally define poisoning attacks in recommendation scenarios from three aspects: attack goal, attack knowledge, and attack capability. For ease of reading, we list key notations in Table \ref{tab: notation} to provide a quick reference guide.

\begin{table}
\caption{Summary of Key Notation}
\begin{tabular}{c|c}
\hline\noalign{\smallskip}
Notation & Definition \\
\noalign{\smallskip}\hline\noalign{\smallskip}
    $n$ & The number of users in the recommender system\\
    $n'$ & The number of poisoning users to the recommender system\\
    $m$ & The number of items in the recommender system\\
    $m'$ & The maximum number of interactive items per poisoning user\\
  $D$ & \makecell{The whole dataset, where each sample $(u,I,R_{u,i})$ denotes that\\ the user $u$s rating on item $i$ is $R_{u,i}$} \\
  $D'$ & The poisoning data in $D$ \\
  $D_U$ & The no rating samples of dataset $D$ \\
  $D_L$ & The training (labeled) data in $D$ \\
  $D_c$ & The initial poisoning data in the proposed attacks \\
  $D_P$ & The generated poisoning data in the proposed attacks \\
  \noalign{\smallskip}\hline
\end{tabular}
\label{tab: notation}
\end{table}

\subsection{Attack goal}
Different shilling attacks may have different intents, but the eventual goal of an attacker may be one of several alternatives. We can divide the attack intents into push attacks, nuke attacks, and random vandalism \cite{si2020shilling}. The push attack (nuke attack) typically aims to increase (decrease) the popularity of the target item. For random vandalism, the attacker combines push attacks and nuke attacks to maximize the recommendation error, making users stop trusting the recommendation model and finally stop using it. We mainly focus on push attacks because nuke attacks can be achieved by increasing the popularity
of non-target items until the target item is not in the user’s recommendation list \cite{yang2017fake}, which in a sense, is equivalent to push attacks. In addition, random vandalism can be seen as a hybrid of push and nuke attacks.
\subsection{Attack knowledge}
According to the attacker’s knowledge, attacks can be divided into high-knowledge attacks, partial-knowledge attacks, and low-knowledge attacks\cite{si2020shilling}. Among them, high-knowledge attacks require the attackers to know detailed knowledge of the target recommender system, such as the algorithm used, specific parameter settings, and even the user’s historical behavior. In a partial-knowledge setting, only part of the user interaction records can be obtained. In addition, the attacker only knows the algorithm of the target model, but the specific training details and model parameters are unknown to the attacker. Low-knowledge attacks further relax knowledge, and the attacker does not even know the algorithm used by the target model. 

In our work, we study the partial-knowledge attack, and low-knowledge attacks will be our future work. Notably, partial-knowledge attacks are operational for attackers. This is because many companies may disclose their recommendation algorithms, such as Amazon\cite{smith2017two} and Netflix\cite{gomez2015netflix}. Besides, the partial data disclosed by users can be easily obtained through web crawlers and other means. In partial-knowledge settings, since the parameters of the target model are unknown, based on the partial data obtained, the attacker trains a local simulator using the same algorithm as the target model but with parameters defined by the attacker. Once the local simulator is trained, the partial-knowledge attack is transferred into a high-knowledge attack for the attacker. Accordingly, the attacker can generate fake users based on the local simulator and inject them into the target model to execute the attack. It needs to be emphasized that benefiting from this transformation, we will introduce our methods based on the high-knowledge attack setting for convenience, but the experiment is indeed based on the partial-knowledge setting.

\subsection{Attack Capability}
The more poisoned data the attacker injects into the recommender system, the more significant the attack performance will be. However, more poisoning data is impractical and will increase the risk of being detected by defense mechanisms \cite{wu2014survey}. In addition, for each fake profile, the more items the attacker interacts with, the more likely to be detected as an anomaly. Based on the above considerations, we limit the number of poisoning profiles and the number of interacting items (filler items) in each profile. Assuming that there are $n$ users and $m$ items in the recommender system, we limit the attacker to registering a maximum of $n'$ malicious users and $m'$ interactive items per user, and $n'\ll n$, $m'\ll m$. We will analyze the different capabilities of attackers in the experiments.

\section{Triple Cooperative Defense}\label{sec4}
\begin{figure}[h]
	\centering
	\includegraphics[width=1\columnwidth]{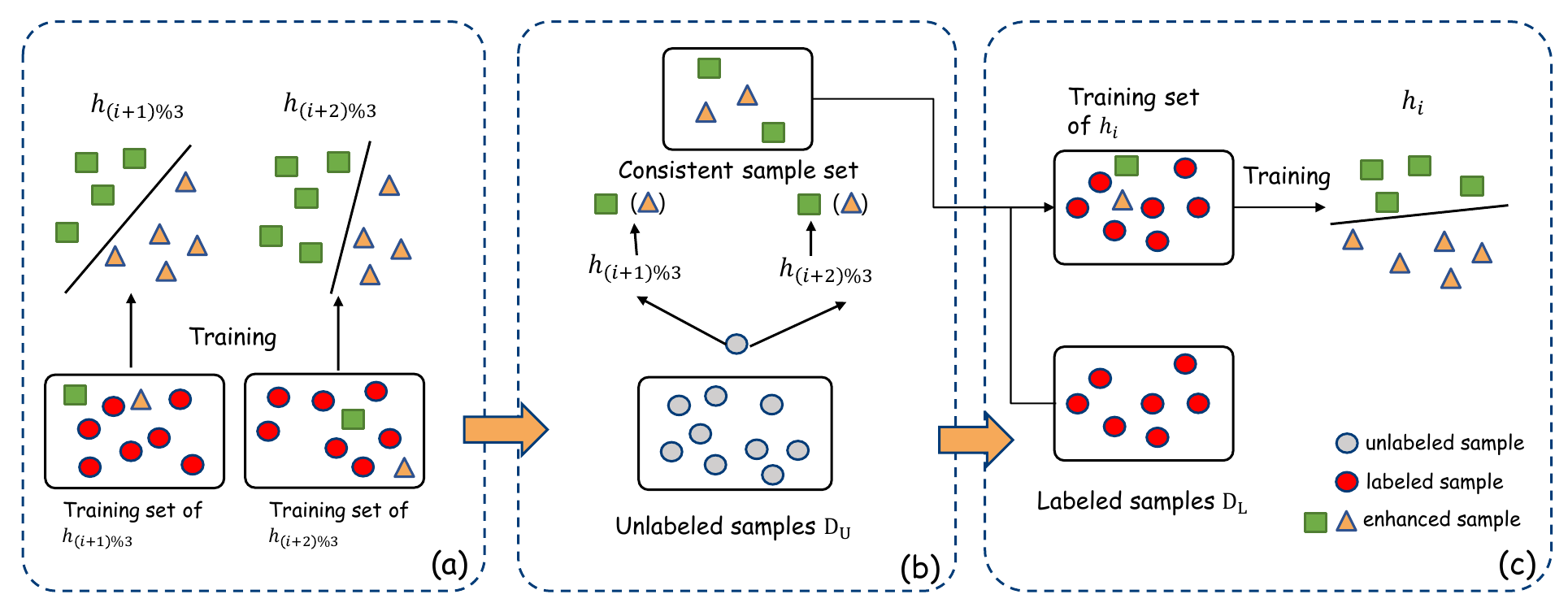}
	\captionsetup{font=scriptsize}
	\caption{The framework of TCD. For model $h_{i}$'s training in each round, (a) the other two models use the same collaborative training; (b) the ratings predicted the same by the other two models are taken as consistent samples; (c) model $h_{i}$ is trained on labeled samples $D_{L}$ and consistent samples.}
	\label{fig:framework}
\end{figure}

In existing work, data-processing-based defense inevitably removes normal data to achieve high recall rates, while the model-based defense is difficult to enjoy both robustness and generalization\cite{zhang2020attacks}. Therefore, it is crucial to design a defense algorithm that maximizes their strengths and circumvents their weaknesses. Recent studies\cite{deldjoo2021survey} demonstrated that robust models require more labeled data\cite{wu2021fight}. Besides, the recommender system is extremely sparse. That is, there is little interactive information about users and items, making a small amount of normal data difficult to support good model training so that it may be misled easily by malicious data and produce biased recommendations. This finding makes us reasonably believe that the vulnerability of the recommendation system is largely due to the lack of data. However, since it takes a lot of manpower and material resources to get labeled data, using a small number of ``expensive'' labeled data is a huge waste of data resources. Considering the above reasons, we constructively propose adding ``cheap'' pseudo ratings with high confidence to improve the recommendation robustness. 

Unfortunately, in the implicit recommendation system concerned in our work, it is challenging to obtain high confidence pseudo scores. This is because the output of recommender systems is prediction scores, not confidence, unlike other areas of machine learning(e.g., in the image field, the output is the prediction probability). So we develop Triple Cooperative Defense (TCD), which uses three models and takes the prediction consistency ratings of any two models as the high confidence pseudo ratings of the remaining model. The recommendation robustness is improved through mutual cooperation among the three. Notably, although more models with majority votes are more beneficial to obtaining high-confidence data, the model training is positively related to the number of models. Therefore, we made a compromise, and we found that the performance of the three models is satisfactory and the training delay is tolerable. The framework is shown in Fig.\ref{fig:framework}. Now we provide details of the proposed TCD for defending against poisoning attacks.

Let $D$ denotes the dataset, $D_{L}$ denotes the rating samples of $D$, where each sample $(u,i,R_{u,i})$ denotes that the user $u$'s rating on item $i$ is $R_{u,i}$, and $D_{U}$ denotes the no rating samples of $D$, where each sample is like $(u,i)$. The goal of the recommendation system $h$ is to predict the accurate rating $\hat{R}_{u,i}=h(u,i)$ of each sample $(u,i)\in D_U$.
In TCD, we extend to three models and denote them as $h_0$, $h_1$, and $h_2$, respectively. For any model,
if the predicted ratings of the other two models are consistent, we have reason to believe that the predicted ratings are high-confident and reliable to be added to the training set to address the difficulty of measuring rating confidence. For instance, if $h_{0}$ and $h_{1}$ agree on the
labeling $\hat{R}_{u,i}$ of $(u,i)$ in $D_{U}$, then $(u,i,\hat{R}_{u,i})$ will be put into the training set for $h_{2}$ as auxiliary training data. It is obvious that in such a scheme, if the prediction of $h_{0}$ and $h_{1}$
on $(u,i)$ is correct, then $h_{2}$ will receive a new sample with high confidence for
further training.
Besides, the predicted ratings are floating points, making judging based on the consistent rating impractical. So we define a projection function $\Pi(\cdot)$ to project continuous ratings onto reasonable discrete ratings. In this way, only when two models give the same rating on $(u,i)$ after projection, we take the rating as the pseudo label and put $(u, i, \Pi(\hat{h_{j}}(u,i)))$ into the $h_{2}$'s training set $D_{L}^{(2)}$.

One explanation for recommendation fragility is that the poisoned data deviates severely from the real data and the sparse (less) real data makes the model learning easy to be dominated by fake data \cite{lin2020attacking}. So intuitively, the augmentation strategy of TCD contributes to magnifying the influence of real profiles and relatively weakening the harm of false profiles.  

\begin{algorithm}
	\caption{Triple Cooperative Defense}
        \label{alg: defense}
	\textbf{Input:}
        The epochs of training $T$, the epochs of pre-training $T_{pre}$, three models $h_{1}(u,i), h_{2}(u,i), h_{3}(u,i)$, labeled data $D_{L}$, unlabeled data $D_{U}$, projection function $\Pi(x)$.
        \begin{algorithmic}[1]
	\FOR{$T_{pre}$ epochs}
		\FOR{$j \in [0,1,2]$}
			\STATE Train $h_{j} $ based on the training set $D_{L}$.
		\ENDFOR
        \ENDFOR
	
	\FOR{$T-T_{pre}$ epochs}
		\FOR{$j \in [0,1,2]$}
			\STATE $D_{L}^{(j)} \gets D_{L}$
			\FOR{every $(u,i) \in D_{U}$}
				\IF{$\Pi(\hat{h}_{(j+1)mod 3}(u,i)) = \Pi(\hat{h}_{(j+2)mod 3}(u,i))$}
				\STATE $D_{L}^{(j)} \gets D_{L}^{(j)} \cup \{(u, i, \Pi(\hat{h}_{(j+1)mod 3}(u,i)))\}$
				\ENDIF
			\ENDFOR
			\STATE Train $h_{j} $ based on training set $D_{L}^{(j)}$
		\ENDFOR
	\ENDFOR
        \end{algorithmic}
\end{algorithm}

The algorithm of TCD is shown in Alg. \ref{alg: defense}. Each model is pre-trained from lines 1 to 5. Then, for each round of training for each model, an unlabeled prediction will be labeled if any two models agree on the labeling, and these pseudo labels with high confidence will be put into the third model's training dataset to reduce the harm that poisoning data do to the model, as shown in lines 7 through 15. After the training, we can perform the recommendation task using any model. In our work, we choose $h_0$ by default. Since the structure of each model is unchanged, the proposed strategy does not have inference delay, which is of more concern to practical applications.  

It is worth noting that in the pre-training phase, we used the same dataset $D_{L}$ for all models. Theoretically, we must choose different training subsets to ensure the model's diversity. This is necessary for other domains, such as the computer version, because the number of parameters in a classifier is independent of the number of samples. However, in recommender systems with extremely sparse data, selecting a subset means that many users are cold-start users, and the parameters of these users cannot be trained, which directly leads to unsatisfactory recommendation performance. Therefore, all label data are selected for pre-training, while different pseudo-labels guarantee the models' diversity in collaborative training. 

\section{Cooperative Training Attack}\label{sec5}
In comparison to heuristic attacks, optimization-based poisoning attacks constitute a more efficacious approach for adapting to a broader spectrum of recommendation patterns \cite{wu2021triple}. However, existing work ignores the bi-level setting of poisoning attacks, which limits the attack performance. In light of this, we revisit poisoning attacks (Section \ref{sec: bi-level}) and propose CoAttack (Section \ref{sec: coattack}), a model that fosters the cooperative optimization of both the model and the attack, as well as GCoAttack (Section \ref{sec: gcoattack}), which further boosts attack by the cooperative optimization of attack and defense mechanisms.

\subsection{Poisoning Attack: A Bi-level Optimization Problem}
\label{sec: bi-level}
Despite researchers' efforts to study optimization-based poisoning attacks, a thorny challenge remains unsolved, namely the Bi-level optimization problem:
\begin{equation}
\label{eq: outer}
\min _{R^{\prime}} \mathcal{L}_{a t k}\left(R, \theta'\right),
\end{equation}
subject to
\begin{equation}
\label{eq: inter}
\theta'=\arg \min _{\theta}\mathcal{L}_{\text {train }}\left(R\cup R', \theta\right).
\end{equation}
Here $R\in\mathbb{R}^{n\times m}$ is the currently observed rating matrix, where
$R_{u,i}$ is the rating given by user $u$ to item $j$, $R^{\prime}\in\mathbb{R}^{n'\times m}$ is the rating matrix
of fake users that we need to solve; $\theta$ ($\theta'$) represents
the parameters of the recommendation model. $\mathcal{L}_{a t k}$ is the attacking objective function, and $\mathcal{L}_{train}$ is the standard training loss (e.g., cross-entropy loss). For the push attack studied in the paper, 
we define $\mathcal{L}_{a t k}$ as follows:
\begin{equation}
\mathcal{L}_{atk}=\sum_{u\in U}\max \left\{\min _{i \in L_u} \log h(u,i)-\log h(u,t),-\kappa\right\},
\end{equation}
where $U$ is the original user set, $L_u$ is the recommendation list for user $u$, $t$ is the target item to be promoted, and $\kappa$ is a positive threshold to make sure that the target item's prediction rating is larger than the $k$-th recommended items. The use of the log operator lessens the dominance effect \cite{huang2021data}. Intuitively, if $\min _{i \in L_u} \log h(u,i)-\log h(u,t)$, then the target item $t$ will be in the top-$k$ recommendation list. Moreover, the
target item $t$ tends to hold a higher rank when the loss $\mathcal{L}_{a t k}$ gets smaller.

However, in a bi-level situation, the model must be retrained after exposure to poisoning attacks. Therefore, the current $\theta$ solution is only an approximation of the optimal solution. Previous works \cite{huang2021data,fang2020influence} have proposed to alternately optimize Formula \ref{eq: outer} and \ref{eq: inter} to approximate the attacking loss better. However, optimizing Formula \ref{eq: inter} is a time-consuming retraining process, which may not be suitable for large-scale recommender systems. Therefore, it is indeed important to efficiently optimize the problem.
\subsection{Co-training Attack}
\label{sec: coattack}
\begin{figure}[h]
	\centering
	\includegraphics[width=0.7\columnwidth]{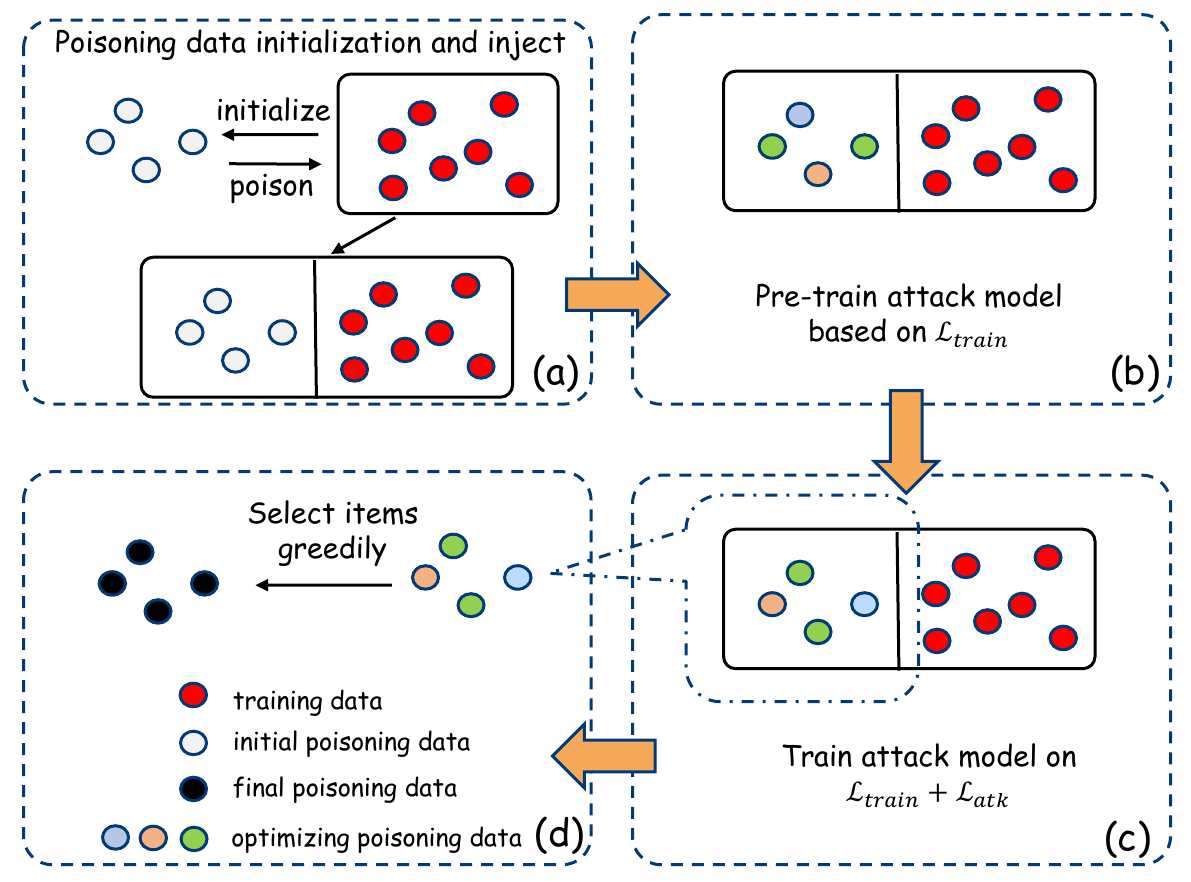}
	\captionsetup{font=scriptsize}
	\caption{The framework of CoAttack. (a) poisoned data initialization and injection. (b) Pre-train attack model on $\mathcal{L}_{train}$. (c) Train attack model on $\mathcal{L}_{train}+\mathcal{L}_{atk}$. (d) Select these items with the highest $m$ ratings in $h(u)$ as $u$'s filler items.}
	\label{fig: coattack}
\end{figure}

An intuitive approach would entail retraining the model (Formula \ref{eq: inter}) for every optimization iteration (Formula \ref{eq: outer}). However, the number of attack optimization instances is equivalent to the number of model retraining, which is computationally infeasible. To this end, Huang et al. \cite{huang2021data} proposed an attack methodology that combines model training and attack optimization. This method fuses attack loss with model training loss, aiming to make the optimized model approximate the ultimate poisoned recommender system, and then the optimized fake profiles are the final poisoning profiles. Taking into account the dynamic changes in model parameters during the poisoning optimization process, this attack circumvents the need for retraining at every optimization step by jointly optimizing the bi-level problem's inner and outer objectives (Formula \ref{eq: outer} and \ref{eq: inter}). 

Nevertheless, it is not without limitations. Firstly, although it obviates the need for model retraining in tandem with the number of optimization, its strategy of generating one user at a time makes the number of model retraining the same as the number of fake users, resulting in time-consuming processes when generating a substantial quantity of fake users. Secondly, this greedy strategy for generating fake users constrains the search space (E.g., when optimizing the $n$-th user, the first $n-1$ users remain fixed), making it challenging to obtain optimal profiles. In response to these limitations, we propose a boosted methodology called Co-training Attack (CoAttack). In CoAttack, we still jointly train the attack optimization and model training; however, the distinction lies in the optimization targeting all candidate poisoning data. This design reduces the number of model training from $n'$ to $1$ and enables us to search within the entirety of the feasible space, which is more conducive to identifying global optima.

As shown in Fig. \ref{fig: coattack}, CoAttack comprises four stages: (1). Poisoning data initialization. We initiate all poisoning profiles by randomly sampling from the distribution of real user ratings, and these profiles will be merged with the real profiles for training. (2). Pre-training stage. We train on data mixed with poisoned data using the standard recommendation training loss $\mathcal{L}_{train}$. After several training epochs, we ensure that the resulting recommendation model possesses the capacity to learn user preferences. Notably, if there were no training stage, the poisoning optimization would be based on a random model, which would be nonsensical. (3) Attack optimization. We incorporate the attacking loss $\mathcal{L}_{atk}$ into the training loss $\mathcal{L}_{train}$ for joint training. Our optimization objective is as follows:
\begin{equation}
    \min_{R',\theta} \mathcal{L}_{atk}(R,\theta)+\mathcal{L}_{train}(R\cup R',\theta).
\end{equation}
Throughout this training process, the model increasingly approximates the attack goal, ultimately culminating in the ideal poisoned model. (4). Fake profile generation. We greedily select the $m'$ items with the highest ratings from the optimized poisoning users as their selected items. Since the optimized ratings are floating-point, we also project these selected item ratings onto reasonable discrete ratings, which serve as final ratings for these items.

\begin{algorithm}
	\caption{Co-training Attack}
 \label{alg: co_attack}
	\textbf{Input:}
        The epochs of training $T$, the epochs of pre-training $T_{pre}$, recommendation model $h(\cdot)$, original data $D$, initial poisoning data $D_{c}$, projection function $\Pi(x)$.
    \begin{algorithmic}[1]
    \STATE Initialize $D_{c}$ according to the distribution of real ratings.
    \STATE $D'=D\cup D_{c}$.
	\FOR{$T_{pre}$ epochs}
		\STATE Train model $h$ based on the dataset $D'$, training loss $\mathcal{L}_{train}$.
	\ENDFOR
	\FOR{$T-T_{pre}$ epochs}
        \STATE Train model $h$ based on the dataset $D'$, training loss $\mathcal{L}_{train}+\mathcal{L}_{atk}$.
	\ENDFOR
    \STATE $D_{p}=\{\}$.
    \FOR{each user $u\in D_{c}$}
        \STATE Get predicted rating vector $h(u)\in\mathbb{R}^m$ for user $u$.
        \STATE Choose these items with the highest $m'$ ratings in $h(u)$ as filler items, and project these ratings to reasonable discrete ratings, denoting as $\hat{R}_u$.
        \STATE $D_{p}=D_{p}\cup \{\hat{R}_u\}$.
    \ENDFOR
    \RETURN poisoning profiles $D_{p}$.
    \end{algorithmic}
\end{algorithm}

The specific algorithm flow is shown in Alg. \ref{alg: co_attack}. Initially, the distribution of real ratings is used to initialize poisoning data $D_{c}$, which is then combined with original data $D$ to create the mixed dataset $D'$, as outlined in lines 1 to 2. Secondly, the model $h$ is pre-trained on the training loss $\mathcal{L}_{train}$ (lines 3 to 5). Thirdly, we train on the combined loss $\mathcal{L}_{train}+\mathcal{L}_{atk}$ for the remaining rounds of attack training (lines 6 through 8). Once joint optimization is complete, we select the top-$m'$ ratings for each poisoning user and project them to reasonable discrete ratings as the final poisoning profile, as shown in lines 9 through 14.

\subsection{Game-based Co-training Attack}
\label{sec: gcoattack}
\begin{figure}[h]
	\centering
	\includegraphics[width=1\columnwidth]{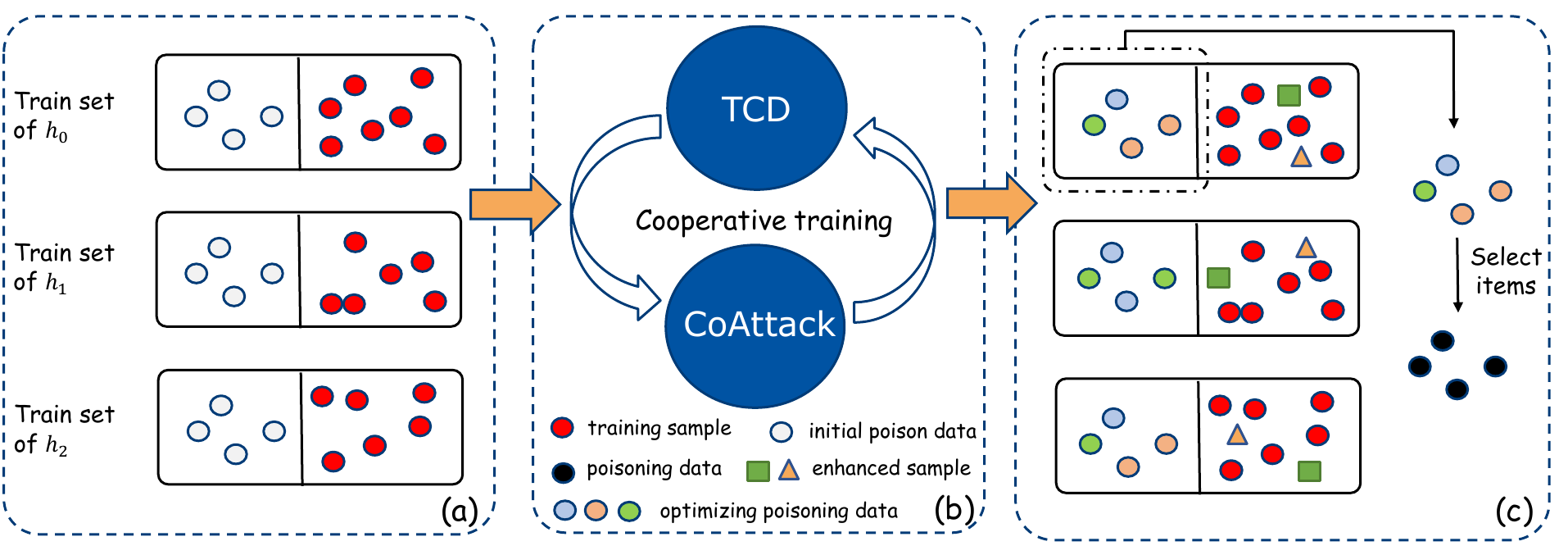}
	\captionsetup{font=scriptsize}
	\caption{The framework of GCoAttack. (a) Pre-train three models on the dataset mixed initial poisoning data. (b) Cooperative train TCD and CoAttack. (c) Choose these items with the highest $m$ ratings in $h_0(u)$ as $u$'s filler items.}
	\label{fig:gcoattack}
\end{figure}

Furthermore, we acknowledge that within the realm of recommender system security, attack and defense inherently constitute an arms race. Effective attacks inevitably become ineffective in subsequent defense. This requires us to revisit attacks. Through a summary of existing attacks, we find that existing attacks are all optimized based on primitive models without any defensive measures. This attack optimized in an overly optimistic environment may be a potential reason for limiting its ability. To address this issue, we characterize the attack-defense dynamic as a game process and conduct joint training among them accordingly. Accordingly, this paper further combines the robust training strategy TCD and the attack strategy CoAttack to propose a Game-based Co-training Attack (GCoAttack).

\begin{algorithm}
	\caption{Game-based Co-training Attack}
    \label{alg: game_co_attack}
	\textbf{Input:}
    The epochs of training $T$, the epochs of pre-training $T_{pre}$, three models $h_{1}(u,i), h_{2}(u,i), h_{3}(u,i)$, labeled data $D_{L}$, unlabeled data $D_{U}$, initial poisoning data $D_c$, projection function $\Pi(x)$
    \begin{algorithmic}[1]
    \STATE Initialize $D_{c}$ according to the distribution of real ratings.
    \STATE $D'=D_{L}\cup D_{U}\cup D_{c}$.
	\FOR{$T_{pre}$ epochs}
		\FOR{$j \in [0,1,2]$}
			\STATE Train $h_{j} $ based on the dataset $D'$, training loss $\mathcal{L}_{train}$.
		\ENDFOR
	\ENDFOR
	
	\FOR{$T-T_{pre}$ epochs}
		\FOR{$j \in [0,1,2]$}
			\STATE $D_{L}^{(j)} \gets D_{L}$.
			\FOR{every $(u,i) \in D_{U}$}
				\IF{$\Pi(\hat{h}_{(j+1)mod 3}(u,i)) = \Pi(\hat{h}_{(j+2)mod 3}(u,i))$}
				\STATE $D_{L}^{(j)} \gets D_{L}^{(j)} \cup \{(u, i, \Pi(\hat{h}_{(j+1)mod 3}(u,i)))\}$
				\ENDIF
			\ENDFOR
			\STATE Train $h_{j} $ based on dataset $D_{L}^{(j)}\cup D'$, training loss $\mathcal{L}_{train}+\mathcal{L}_{atk}$.
		\ENDFOR
	\ENDFOR
    \STATE $D_{p}=\{\}$.
    \FOR{each user $u\in D_{c}$}
        \STATE Get predicted rating vector $h_0(u)\in\mathbb{R}^m$ for user $u$.
        \STATE Choose these items with the highest $m'$ ratings in $h(u)$ as filler items, and project these ratings to reasonable discrete ratings, denoting as $\hat{R}_u$.
        \STATE $D_{p}=D_{p}\cup \{\hat{R}_u\}$.
	\ENDFOR
    \RETURN poisoning profiles $D_{p}$.
    \end{algorithmic}
\end{algorithm}

To better understand the dynamics of this game, we can envision it as a competition between two players: the attacker and the recommender system's defense mechanism. The attacker's goal is to inject poisoning profiles, tricking the recommender system into promoting specific target items to a larger audience. Meanwhile, the defense mechanism aims to develop a robust recommendation model that accurately captures users' genuine preferences. This situation can be viewed as a mutually competitive zero-sum game, where the success of one side comes at the expense of the other. Consequently, we can expect that attacks conceived in this challenging environment will exhibit the highest attack potential.

In the GCoAttack framework, we have two key players: the attacker, represented by CoAttack, and the defender, which corresponds to TCD. As previously discussed, TCD aims to bolster recommendation system robustness by collaboratively training three models. In contrast, CoAttack focuses on efficient attacks by cooperatively training the attacker and the model, essentially creating a zero-sum game dynamic between them.
Here's how GCoAttack operates: Initially, we initialize the poisoned users and pre-train the three TCD models using standard recommendation loss $\mathcal{L}_{train}$. In each round of attack optimization, TCD generates high-confidence pseudo-labels, contributing to training and enhancing robustness. Subsequently, CoAttack optimizes the fake users based on augmented data, considering both attack loss and training loss simultaneously. CoAttack, in essence, attempts to breach TCD's defenses within this alternating optimization process. It strives to identify the optimal attack strategy to maximize attack potential. The framework is shown in Fig. \ref{fig:gcoattack}.

The detailed algorithm of GCoAttack is shown in Alg. \ref{alg: game_co_attack}. We initialize the poisoning data $D_{c}$ and inject them into the recommender system from lines 1 to 2. During the pre-training phase, each model is trained based on the standard loss $\mathcal{L}_{train}$. Subsequently, for each round of attack optimization, high-confidence pseudo-labels are generated for every model and injected into the dataset, as illustrated in lines 10-14. GCoAttack then optimizes the attack based on the loss $\mathcal{L}_{train}+\mathcal{L}_{atk}$ using the augmented dataset, as depicted in line 16. Upon the completion of attack optimization, the final poisoning users are selected using a greedy strategy similar to CoAttack, as shown in lines 20-24. It is worth noting that, in our experiments, we discovered that the attacking quality derived from any model $h_i$ is relatively equivalent; thus, by default, we opt for model $h_0$.

\section{EXPERIMENT}\label{sec6}

\subsection{Experimental Settings}
\subsubsection{Datasets}
We use three real-world datasets commonly used
in the security studies \cite{christakopoulou2019adversarial,yuan2019adversarial} of the recommender system, including FilmTrust\footnote{https://www.librec.net/datasets/flmtrust.zip}, ML-100K\footnote{https://grouplens.org/datasets/movielens}
(MovieLens-100K), and ML-1M\footnote{https://grouplens.org/datasets/movielens}(MovieLens-1M). ML-100K includes 943 users who
have rated 1,682 movies for 100,000 ratings. ML-1M comprises 6,040
users who have rated 3,706 movies about one million times. For
FilmTrust, the same pretreatment as \cite{lin2020attacking} is used to filter cold-start
users who seriously affect the recommender system (the rating
number is less than 15), leaving 796 users with trust ratings for
2011 movies. Table \ref{tab: statistics} lists the detailed statistics of these datasets. All
ratings are from 1 to 5, and we normalized them to [0, 1] in the
experiments. For each dataset, we randomly select a positive sample
from each user for testing, and the rest are used as the training set
and verification set in a 9:1 ratio.

\begin{table}[htbp]
	\centering
	\caption{Statistics of datasets}
	\setlength{\tabcolsep}{5mm}{
	\begin{tabular}{ccccc}
		\toprule  
		Dataset&users&items&ratings&sparsity \\ 
		\midrule  
		FilmTrust&796&2011&30880&98.07\\
		ML-100K&943&1682&100000&93.70\\
		ML-1M&6040&3706&1000209&95.53\\
		\bottomrule  
	\end{tabular}}
 \label{tab: statistics}
\end{table}

\subsubsection{Attack Methods}
Here we use the following attacks for robustness validation:
\begin{itemize}
	\item \textbf{Random Attack} \cite{lam2004shilling}: This attack assigns the maximum rating
	to the target item and rates selected items according to the normal
	distribution of all user ratings at random.
	
	\item \textbf{Average Attack} \cite{lam2004shilling}: The only difference from Random Attack
	is that the non-target selected item is randomly rated with the
	normal rating distribution of items.
	
	\item \textbf{AUSH Attack} \cite{lin2020attacking}: This attack uses GAN to generate fake
	users to carry out attacks imperceptibly and assigns the highest
	rating to the target item.
	
	\item \textbf{PGA Attack} \cite{li2016data}: This attack builds an attack objective and uses
	SGD to update the poisoned user’s ratings to optimize the objective.
	Finally, the first items with the largest ratings are selected as
	the fake user’s filler items.
	
	\item \textbf{TNA Attack} \cite{fang2020influence}: This attack selects a subset of the most influential users in the dataset and optimizes the rating gap between the
	target item and top-K items in the user subset. Here we use S-TNA. 

 \item \textbf{DL Attack} \cite{huang2021data}: The attack problem of non-convex integer programming is solved by multiple approximations, and
traditional training and poisoning training are combined to generate fake users. Notably, CoAttack is inspired by it, and we will compare them in Seciton \ref{sec: attack_comparision} to verify the effectiveness of the proposed attack.
\end{itemize}

In the context of the partial-knowledge attacks examined in this paper, the attacker leverages captured partial data to reconstruct a local simulator that closely resembles the target model. Subsequently, the attacker employs this local simulator as a white-box resource for conducting attacks. The transferability of these attacks ensures their potential harm.

\subsubsection{Defense Methods}
We compare the proposed TCD with the following robust algorithms:
\begin{itemize}
	\item \textbf{Adversarial Training(AT)}\cite{he2017neural}: In each training step, it first uses SGD to optimize the inner objective to generate small perturbations, adds them to the parameters, and then performs training.
	
	\item \textbf{Random Adversarial Training(RAT)}\cite{he2017neural}: In each training step, it first uses the truncated normal distribution $\mathcal{N}(0,0.01)$ to generate small perturbations, adds them to the parameters, and then performs training.
	
\end{itemize}
These methods face a trade-off between generalization and robustness. Greater noise improves robustness but significantly reduces generalization. As a compromise, we've set the maximum noise value to 0.03.
\subsubsection{Evaluation Metric}
We first use HR (Hit Ratio), just like \cite{wu2021fight}, which
calculates the proportion of test items that appear in the user’s top-K recommendation list. Setting a large K helps make apparent comparisons between defense methods and collaborative filtering is often used for candidate selection in practical recommendations, so it is more instructive to select a larger K to ensure a high recall \cite{he2018adversarial}, and we set K to 50 in the experiments. Besides, we use robustness improvement
$RI = 1- (HR_{defense} - HR_{orgin})/(HR_{attack} - HR_{orgin})$ defined in \cite{wu2021fight}. A value closer to 1 indicates better robustness.    Finally, we introduce the Rank Shift metric, which quantifies the difference between the rank of the targeted item before and after the attack. A larger deviation from 0 signifies a more significant impact of the attack.
Our reported results are based on the averages from 30 repeated independent experiments. We also conduct paired t-tests when necessary to assess statistical significance.

\subsubsection{Parameters Setting}
We concern with the MF-based collaborative filtering method, and we set the latent factor
dimension $d$ to 128, the batch size to 2048, and the regularization parameter to 0.005. During the training phase, the model undergoes training for 40 epochs, utilizing the Adam optimizer for optimization. The final model selection is based on achieving the smallest Mean Squared Error (MSE). For the partial-knowledge attack studied in our work, unless otherwise specified, we set the data obtained by the attacker as 40\%, the attack size as 3\%, and the number of filler items as the average number of real users. Importantly, this is not in conflict with the condition where $m' << m$, as the average number of user ratings is significantly smaller than the total number of items in the dataset, e.g., the number of filler items in Yelp is 38, which is far less than the total number of 25,602 items.

For the proposed TCD, the pre-training epoch $Tpre$ is set to 1, 4, and 2 in FilmTrust, ML-100K, and ML-1M, respectively. For the number of pseudo-labels used, for the smaller FilmTrust and ML-100K, we use all high-confidence pseudo-labels, while for the larger ML-1M, we randomly select 20\% for model training efficiency (comparison of other ratios can be found in Section \ref{sec: label ratio}). For the proposed CoAttack and GCoAttack, the pre-training epoch $T_{pre}$ is set to 1, and the threshold $\kappa$ is set to 0.2. Besides, the ratio of high-confidence pseudo-labels is set to 100\%, 100\%, and 20\%, similar to TCD.

For the target items of attacks, we learn two types: (1) random items: randomly selected from all items, and (2) unpopular items: randomly selected from items with the number of ratings less than 5. For both types of items, we choose 5 items as target items. If you wish to access the source code for our work, it is available at the following URL:  \url{https://github.com/greensun0830/Cotraining-Attack}.

\subsection{Result Analysis regarding Attack}

\begin{table*}[htbp]
	\centering
	\caption{Attack performance (HR@50) under different attack knowledge-cost. *, **, and *** indicate that the improvements over the best baseline are statistically significant for $ p<0.05, p<0.01 $, and $ p<0.001 $, respectively.}
	\begin{adjustbox}{width=1\textwidth}
		\small
		\begin{tabular}{c|c|c|c|c|c|c|c|c|c|c|c|c|c}
			\noalign{\smallskip} \hline \noalign{\smallskip}
			\multirow{3}[3]{*}{Dataset} & \multicolumn{7}{c|}{Random Items}                      & \multicolumn{6}{c}{Unpopular Items} \\
			\noalign{\smallskip} \cmidrule{2-14} \noalign{\smallskip}
			& \multirow{2}[2]{*}{Attack} &\multirow{2}[2]{*}{Origin} & \multicolumn{5}{c|}{Attack Knowledge-cost}            & \multirow{2}[2]{*}{Origin} & \multicolumn{5}{c}{Attack Knowledge-cost} \\
			\noalign{\smallskip} \cmidrule{4-8}  \cmidrule{10-14}\noalign{\smallskip} 
			&       &       & 0.2   & 0.4   & 0.6   & 0.8   & 1     &       & 0.2   & 0.4   & 0.6   & 0.8   & 1 \\
			\noalign{\smallskip} \hline \noalign{\smallskip}
			\multirow{9}[0]{*}{Filmtrust} & Average & 0.2065 & 0.1165 & 0.1210 & 0.1350 & 0.1461 & 0.1431 & 0.0000 & 0.0028 & 0.0020 & 0.0024 & 0.0020 & 0.0029 \\
			& Random & 0.2065 & 0.1596 & 0.1511 & 0.1491 & 0.1449 & 0.1564 & 0.0000 & 0.0046 & 0.0029 & 0.0030 & 0.0026 & 0.0036 \\
			& AUSH  & 0.2065 & 0.1473 & 0.1807 & 0.2944 & 0.3597 & 0.3668 & 0.0000 & 0.0384 & 0.0363 & 0.0617 & 0.0562 & 0.0921 \\
			& PGA   & 0.2065 & 0.1106 & 0.1250 & 0.1453 & 0.1753 & 0.1817 & 0.0000 & 0.0019 & 0.0019 & 0.0051 & 0.0039 & 0.0102 \\
			& TNA   & 0.2065 & \underline{0.7299} & \underline{0.6826} & 0.5736 & \underline{0.6619} & 0.4762 & 0.0000 & \underline{0.5126} & \underline{0.6602} & 0.3423 & 0.1728 & 0.0996 \\
   & DL & 0.2065 & 0.3825 & 0.5187 & 0.5787 & 0.6001 & 0.5412 & 0.0000 & 0.0407 & 0.0603 & 0.0812 & 0.0611 & 0.1037 \\
              & CoAttack & 0.2065 & 0.5678 & 0.6443 & \underline{0.7334} & 0.5646 & \underline{0.7383} & 0.0000 & 0.1117 & 0.4429 & \underline{0.4430} & \underline{0.3521} & \underline{0.3140} \\
			& GCoAttack  & 0.2065 & \textbf{0.8115} & \textbf{0.8578} & \textbf{0.8814} & \textbf{0.8818} & \textbf{0.8843} & 0.0000 & \textbf{0.7405} & \textbf{0.7263} & \textbf{0.8346} & \textbf{0.7830} & \textbf{0.8711} \\
   \cmidrule{2-14}
   & p-value & & ***   & ***   & ***   & ***   & ***  & & ***   &    & *** & ***   & ***\\
			\noalign{\smallskip} \hline\hline \noalign{\smallskip}
			\multirow{9}[0]{*}{ML-100K} & Average & 0.0535 & 0.1373 & 0.2045 & 0.2691 & 0.2764 & 0.2669 & 0.0000 & 0.0389 & 0.2098 & 0.6832 & 0.6144 & 0.5917 \\
			& Random & 0.0535 & 0.0860 & 0.1127 & 0.1116 & 0.1203 & 0.1291 & 0.0000 & 0.1262 & 0.1341 & 0.2138 & 0.1407 & 0.1884 \\
			& AUSH  & 0.0535 & 0.1946 & 0.3336 & 0.3829 & 0.3928 & 0.3801 & 0.0000 & 0.0539 & 0.2818 & 0.7997 & 0.8310 & 0.8743 \\
			& PGA   & 0.0535 & 0.2241 & 0.2043 & 0.1642 & 0.1930 & 0.1812 & 0.0000 & 0.3894 & 0.4153 & 0.3972 & 0.3442 & 0.2932 \\
			& TNA   & 0.0535 & 0.1574 & 0.3450 & 0.3452 & 0.3155 & 0.3485 & 0.0000 & 0.7872 & 0.4146 & 0.7462 & 0.7383 & 0.7171 \\
   & DL & 0.0535 & 0.2543 & 0.3952 & 0.3637 & 0.4042 & 0.4534 & 0.0000 & 0.9223 & 0.8299 & 0.8728 & 0.8309 & 0.9094 \\
   & CoAttack  & 0.0535 & \underline{0.3308} & \underline{0.4280} & \underline{0.4707} & \underline{0.4327} & \underline{0.4869} & 0.0000 & \underline{0.9260} & \underline{0.8976} & \underline{0.9407} & \underline{0.8592} & \underline{0.8969} \\
			& GCoAttack & 0.0535 & \textbf{0.3551} & \textbf{0.5008} & \textbf{0.5834} & \textbf{0.5978} & \textbf{0.5925} & 0.0000 & \textbf{0.9903} & \textbf{0.9874} & \textbf{0.9898} & \textbf{0.9906} & \textbf{0.9884} \\
   \cmidrule{2-14}
    & p-value & & ***   & ***   & ***   & ***   & ***  & & ***   & ***   & *** & ***   & ***\\
			\noalign{\smallskip} \hline\hline \noalign{\smallskip}
			\multirow{9}[0]{*}{ML-1M} & Average & 0.0000 & 0.2052 & 0.2729 & 0.3123 & 0.3119 & 0.3392 & 0.0000 & 0.9317 & 0.9490 & 0.9511 & 0.9557 & 0.9536 \\
			& Random & 0.0000 & 0.0609 & 0.0694 & 0.0687 & 0.0626 & 0.0725 & 0.0000 & 0.7731 & 0.7799 & 0.7631 & 0.7472 & 0.7934 \\
			& AUSH  & 0.0000 & 0.2255 & \textbf{0.3171} & \textbf{0.3304} & \textbf{0.3632} & \textbf{0.3753} & 0.0000 & 0.9768 & 0.9805 & 0.9761 & 0.9819 & 0.9863 \\
			& PGA    & 0.0000 & 0.0986 & 0.1150 & 0.1003 & 0.0515 & 0.0446 & 0.0000 & 0.9693 & 0.9515 & 0.9403 & 0.9260 & 0.9297 \\
			& TNA   & 0.0000 & 0.0665 & 0.2913 & 0.3288 & 0.3274 & 0.3283 & 0.0000 & 0.9267 & 0.9489 & 0.9535 & 0.9606 & 0.9554 \\
   & DL & 0.0000 & 0.2236 & 0.2475 & 0.2764 & 0.2748 & 0.2680 & 0.0000 & 0.9810 & 0.9736 & 0.9662 & 0.9754 & 0.9650 \\
   & CoAttack & 0.0000 & \underline{0.2261} & 0.2478 & 0.2436 & 0.2263 & 0.2344 & 0.0000 & \underline{0.9813} & \underline{0.9898} & \underline{0.9918} & \underline{0.9902} & \underline{0.9951} \\
			& GCoAttack & 0.0000 & \textbf{0.2662} & \underline{0.3002} & \underline{0.3293} & \underline{0.3359} & \underline{0.3356} & 0.0000 & \textbf{0.9981} & \textbf{0.9978} & \textbf{0.9980} & \textbf{0.9979} & \textbf{0.9975} \\
   \cmidrule{2-14}
    & p-value & & ***   &    &    &    &   & & ***   & ***   & *** & ***   & ***\\
			\noalign{\smallskip} \hline \noalign{\smallskip}
		\end{tabular}%
	\end{adjustbox}
	\label{tab: attack_performance}
\end{table*}%

\subsubsection{Performance Comparison}
\label{sec: attack_comparision}
This section compares the proposed attacks with the existing state-of-the-art attack methods. Table \ref{tab: attack_performance} illustrates HR@50 of target items under varying degrees of attack knowledge. 
Firstly, the proposed attacks (CoAttack and GCoAttack) significantly outperform the baselines in most scenarios, such as attacking unpopular items on FilmTrust, where the average attack improvement reaches an astounding 258\%. This demonstrates the rationality of considering bi-level optimization in poisoning attacks. Secondly, it can be observed that as the attack knowledge increases, the performance of various attacks exhibits an upward trend. This is expected, as the attackers can better understand the true data distribution and tailor their poisoning efforts for more users. Lastly, model-based optimization attacks (e.g., TNA, DL) are superior to heuristic attacks (e.g., Average attack, Random attack). This validates our earlier discussion that heuristic attack methods, which solely rely on generating fake profiles based on general experience, cannot adapt to all recommendation patterns and therefore fail to achieve satisfactory attack performance. Even these heuristic attacks reduce the exposure rate of target items, which emphasizes the significance of studying optimization-based attacks.

\begin{figure*}
\centering 
{\includegraphics[width=0.48\textwidth]{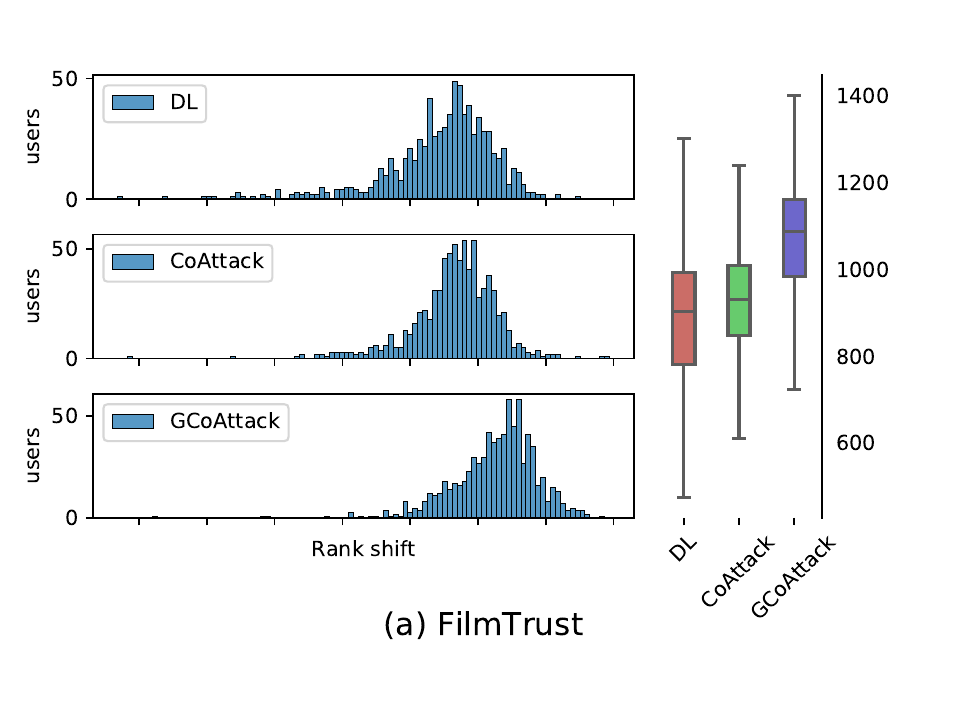}\quad}
{\includegraphics[width=0.48\textwidth]{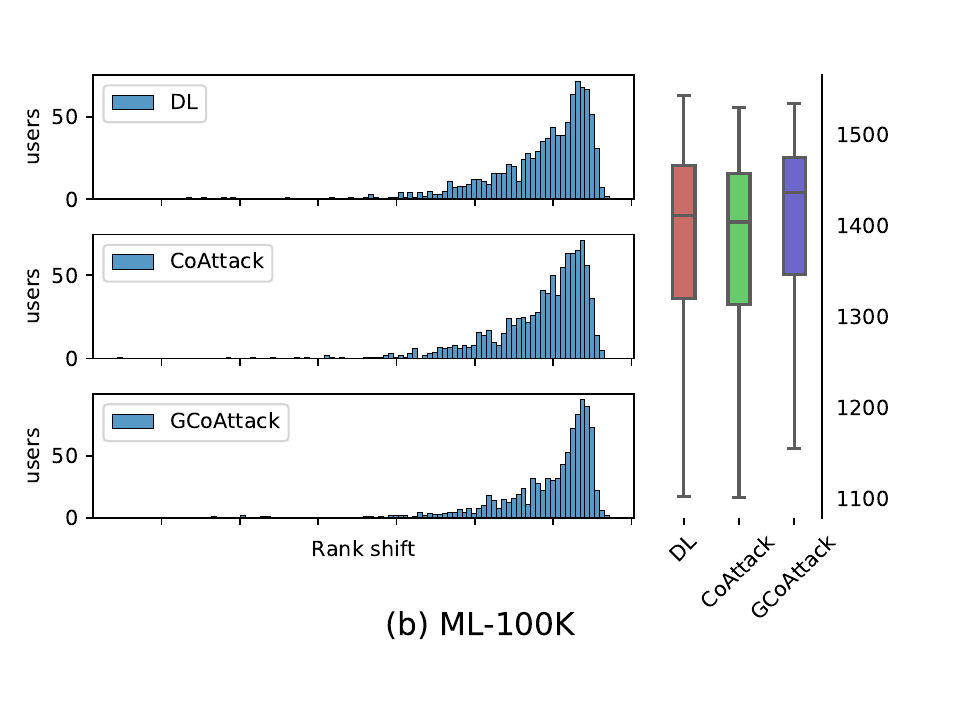}\par\medskip}
{\includegraphics[width=0.48\textwidth]{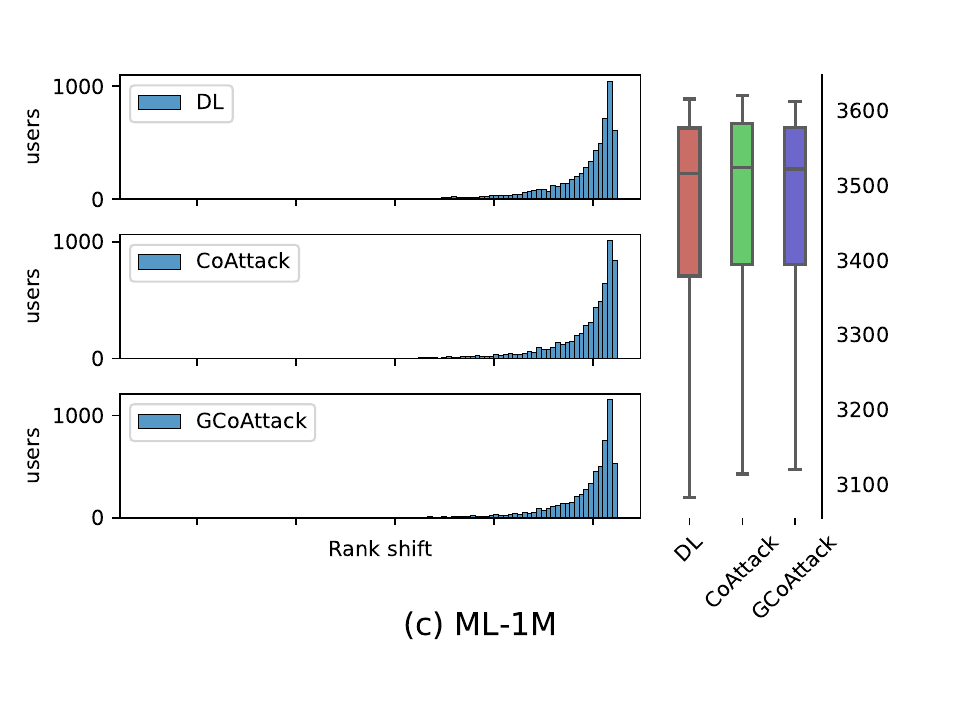}}
\caption{Rank shift distribution of target items (unpopular items). The greater the rank shift, the more harmful the attack.}
  \label{fig: attack rank}
\end{figure*}

In addition, we conduct a comparative analysis of the proposed CoAttack, and GCoAttack with DL (CoAttack is inspired by it) to further verify the effectiveness of our designs. Firstly, the comparison between CoAttack and DL in Table \ref{tab: attack_performance} shows that the attack performance of CoAttack using all poison user optimization is significantly improved compared to DL using single user optimization. This validates that the larger search space in CoAttack facilitates the discovery of optimal poisoning profiles. Secondly, when comparing CoAttack with GCoAttack, we notice a further improvement in attack performance, underscoring the importance of optimizing attacks in game-based settings. That is, more stringent environments give rise to more potent attacks. In addition, to compare the three models more intuitively, we plot the shifting distribution of recommended ranking of the target items after the attack, as depicted in Fig. \ref{fig: attack rank}. A larger rank shift signifies superior attack performance. It is apparent that the performance of the three methods progressively improves, further corroborating the rationality of employing cooperative training based on all fake profiles (i.e., CoAttack) and game-theoretic cooperative training (i.e., GCoAttack).

\subsubsection{Performance under Different Attack Sizes}
\begin{figure}[h]
	\centering
	\includegraphics[width=1.\columnwidth]{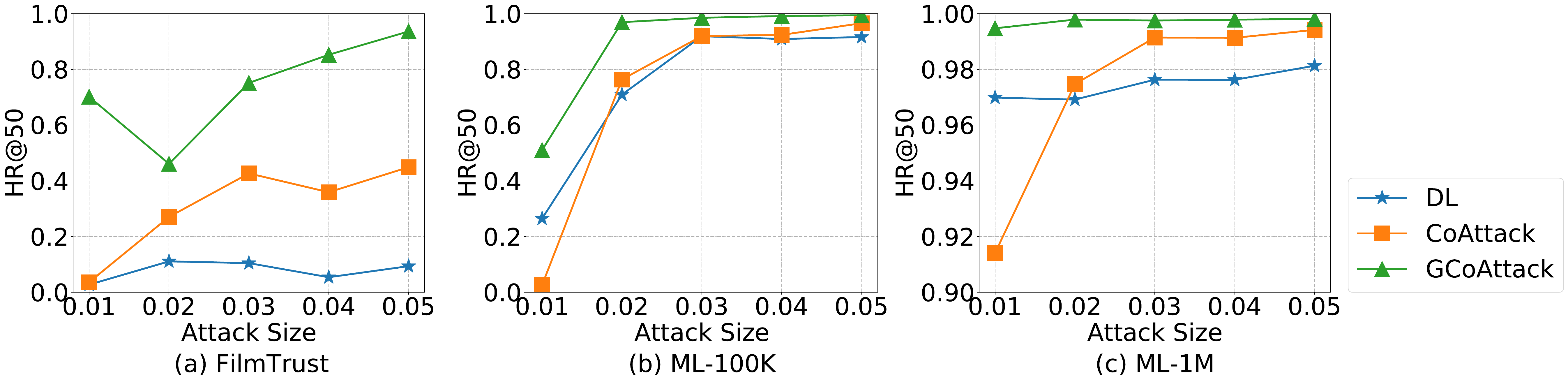}
	\caption{Attack performance regarding random items under different attack sizes.}
	\label{fig:attack size random}
\end{figure}

\begin{figure}[h]
	\centering
	\includegraphics[width=1.\columnwidth]{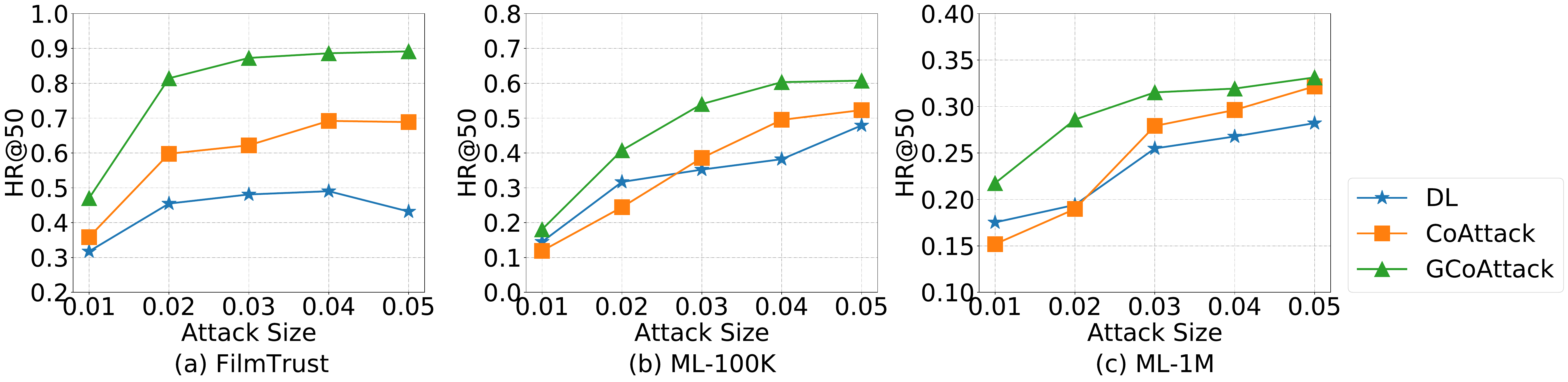}
	\caption{Attack performance regarding unpopular items under different attack sizes.}
	\label{fig:attack size unpopular}
\end{figure}

Theoretically, any recommendation system is inherently susceptible to adversarial manipulation without constraining the number of poisoning profiles. Nonetheless, increased poisoning data entails a heightened probability of detection. Considering these, we investigate attack performance under various poisoning sizes, as illustrated in Fig. \ref{fig:attack size random} and \ref{fig:attack size unpopular}. Firstly, as anticipated, the intensity of the attack is positively correlated with the number of poisoning instances. Secondly, under different amounts of poison data settings, CoAttack consistently outperforms DL, while GCoAttack surpasses CoAttack. It further substantiates the effectiveness of the two devised attack strategies from the sensitivity to attack size perspective.

\begin{table*}[htbp]
  \centering
  \small
  \caption{The performance in target items (robustness). *, **, and *** indicate that the improvements over the best results of baselines are statistically significant for $ p<0.05, p<0.01 $, and $ p<0.001 $, respectively.}
  \renewcommand\arraystretch{0.85}
  \begin{adjustbox}{width=1\textwidth}
    \begin{tabular}{c|c|cccccccc}
    \hline \noalign{\smallskip}
    \multirow{2}[0]{*}{Dataset} & \multirow{2}[0]{*}{Method} & \multicolumn{8}{c}{Random Items} \\
    \cmidrule{3-10}
          &       & Average & Random & AUSH  & PGA   & TNA   & DL    & CoAttack & GCoAttack \\
    \noalign{\smallskip} \hline \noalign{\smallskip}
    \multirow{5}[0]{*}{FilmTrust} & Origin & 0.2065 & 0.2065 & 0.2065 & 0.2065 & 0.2065 & 0.2065 & 0.2065 & 0.2065 \\
    & Attack    & 0.1210 & 0.1511 & 0.1764 & 0.1807 & 0.1250 & 0.6826 & 0.6443 & 0.8578  \\
          & AT    & 0.1119 & \textbf{0.1672} & 0.1764 & 0.1276 & 0.6088 & 0.5222 & 0.6039 & 0.8362  \\
          & RAT  & 0.1143 & 0.1479 & \textbf{0.1773} & 0.1244 & 0.6806 & 0.5055 & 0.6032 & 0.8572  \\
          & TCD   & \textbf{0.1188} & 0.1224 & 0.2386 & \textbf{0.1287} & \textbf{0.5125} & \textbf{0.1810} & \textbf{0.1693} & \textbf{0.1721} \\
          \cmidrule{2-10}
          & p-value & **    &       &     & ***   & ***   & ***   & ***   & *** \\
          \midrule
    \multirow{5}[0]{*}{ML-100K} & Origin & 0.0535 & 0.0535 & 0.0535 & 0.0535 & 0.0535 & 0.0535 & 0.0535 & 0.0535  \\
    & Attack    & 0.2045 & 0.1127 & 0.3336 & 0.2043 & 0.3450 & 0.3952 & 0.4280 & 0.5008  \\
          & AT    & 0.2088 & 0.1157 & 0.3387 & 0.2264 & 0.3148 & 0.3855 & 0.3356 & 0.4782 \\
          & RAT   & 0.1984 & 0.1144 & 0.3385 & 0.1846 & 0.2841 & 0.3999 & 0.3888 & 0.5333  \\
          & TCD   & \textbf{0.0544} & \textbf{0.0495} & \textbf{0.0528} & \textbf{0.0544} & \textbf{0.0489} & \textbf{0.0942} & \textbf{0.0981} & \textbf{0.1026} \\
          \cmidrule{2-10}
          & p-value & ***   & ***   & ***   & ***   & ***   & ***   & ***   & *** \\
          \midrule
    \multirow{5}[0]{*}{ML-1M} & Origin & 0.0000 & 0.0000 & 0.0000 & 0.0000 & 0.0000 & 0.0000 & 0.0000 & 0.0000  \\
    & Attack    & 0.2729 & 0.0694 & 0.2255 & 0.0986 & 0.0665 & 0.2236 & 0.2261 & 0.2662  \\
          & AT    & 0.1018 & 0.0479 & 0.0602 & 0.0700 & 0.1183 & 0.1518 & 0.1623 & 0.1841  \\
          & RAT   & 0.2518 & 0.0590 & 0.2770 & 0.1153 & 0.2649 & 0.2427 & 0.2380 & 0.2901  \\
          & TCD   & \textbf{0.0061} & \textbf{0.0066} & \textbf{0.0047} & \textbf{0.0258} & \textbf{0.0051} & \textbf{0.0835} & \textbf{0.3588} & \textbf{0.3662}  \\
          \cmidrule{2-10}
          & p-value & ***   & ***   & ***   & ***   & ***   & ***   & ***   & *** \\
          \midrule
          \midrule
    \multirow{2}[0]{*}{Dataset} & \multirow{2}[0]{*}{Method} & \multicolumn{8}{c}{Unpopular Items} \\
    \cmidrule{3-10}
          &       & Average & Random & AUSH  & PGA   & TNA   & DL    & COATK & GCOATK \\
    \midrule
    \multirow{5}[0]{*}{FilmTrust} & Origin & 0.0000  & 0.0000  & 0.0000  & 0.0000  & 0.0000  & 0.0000  & 0.0000  & 0.0000  \\
    & Attack    & 0.0020 & 0.0029 & 0.0363 & 0.0019 & 0.6602 & 0.0603 & 0.4429 & 0.7263  \\
          & AT    & 0.0016 & 0.0027 & 0.0483 & 0.0020 & 0.6046 & 0.0995 & 0.4177 & 0.8051  \\
          & RAT   & 0.0017 & 0.0027 & 0.0586 & 0.0018 & 0.5890 & 0.0800 & 0.4467 & 0.7708   \\
          & TCD   & \textbf{0.0008} & \textbf{0.0016} & \textbf{0.0046} & \textbf{0.0007} & \textbf{0.0623} & \textbf{0.0133} & \textbf{0.0122} & \textbf{0.0362} \\
          \cmidrule{2-10}
          & p-value & ***   & ***   & ***   & ***   & ***   & ***   & ***   & *** \\
          \midrule
    \multirow{5}[0]{*}{ML-100K} & Origin & 0.0000  & 0.0000  & 0.0000  & 0.0000  & 0.0000  & 0.0000  & 0.0000  & 0.0000  \\
    & Attack    & 0.2098 & 0.1341 & 0.2818 & 0.4153 & 0.4146 & 0.8299 & 0.9876 & 0.9874  \\
          & AT    & 0.3051 & 0.1580 & 0.2338 & 0.5953 & 0.5151 & 0.8861 & 0.9309 & 0.9820  \\
          & RAT  & 0.1957 & 0.1522 & 0.2450 & 0.4472 & 0.4129 & 0.8645 & 0.9291 & 0.9856  \\
          & TCD  &\textbf{0.0010} & \textbf{0.0013} & \textbf{0.0010} & \textbf{0.0015} & \textbf{0.0019} & \textbf{0.0032} & \textbf{0.0012} & \textbf{0.0069}\\
          \cmidrule{2-10}
          & p-value & ***   & ***   & ***   & ***   & ***   & ***   & ***   & *** \\
          \midrule
    \multirow{5}[0]{*}{ML-1M} & Origin & 0.0000  & 0.0000  & 0.0000  & 0.0000  & 0.0000  & 0.0000  & 0.0000  & 0.0000  \\
    & Attack    & 0.9490 & 0.7799 & 0.9805 & 0.9515 & 0.9489 & 0.9736 & 0.9898 & 0.9978  \\
          & AT   & 0.9553 & 0.6698 & 0.9636 & 0.9458 & 0.9502 & 0.9729 & 0.9926 & 0.9965  \\
          & RAT   & 0.9504 & 0.7492 & 0.9802 & 0.9523 & 0.9441 & 0.9760 & 0.9894 & 0.9978   \\
          & TCD   &\textbf{0.0321} & \textbf{0.0258} & \textbf{0.0344} & \textbf{0.0301} & \textbf{0.0288} & \textbf{0.1444} & \textbf{0.8046} & \textbf{0.8644}\\
          \cmidrule{2-10}
          & p-value & ***   & ***   & ***   & ***   & ***   & ***   & ***   & *** \\
          \midrule
    \end{tabular}%
    \end{adjustbox}
  \label{tab:robustness}%
\end{table*}%

\subsection{Result Analysis regarding Defense}

\subsubsection{Robustness}

In this evaluation, we assess the mitigating effect of various defense methods on the hit ratio (HR) of target items, as shown in Table \ref{tab:robustness}. Here, ``Origin" refers to the unperturbed model, while ``Attack" denotes the attacked victim model subjected to various attacks. First, in most cases, the employed defense methods show a positive effect in weakening the attack's damage with respect to HR. Second, our proposed TCD stands out by achieving impressive defense results, almost matching the performance of the unperturbed model. On average, TCD reduces the impact of attacks on random items by over 88\% and unpopular items by over 82\%, significantly outperforming baseline defenses. Third, when attacking FilmTrust's unpopular items, the performance of TCD against Random and Average is slightly inferior compared to the defense against other attacks. In contrast, almost every performance of TCD on ML-100k and ML-1M is better than that of baselines. We suspect that the smaller size of the FilmTrust dataset may not adequately represent real data, making it easier for adversarial training to identify and learn non-robust features of adversarial data. This also presents a more formidable challenge for TCD in detecting such non-robust features.

\begin{figure*}
\centering 
{\includegraphics[width=0.48\textwidth]{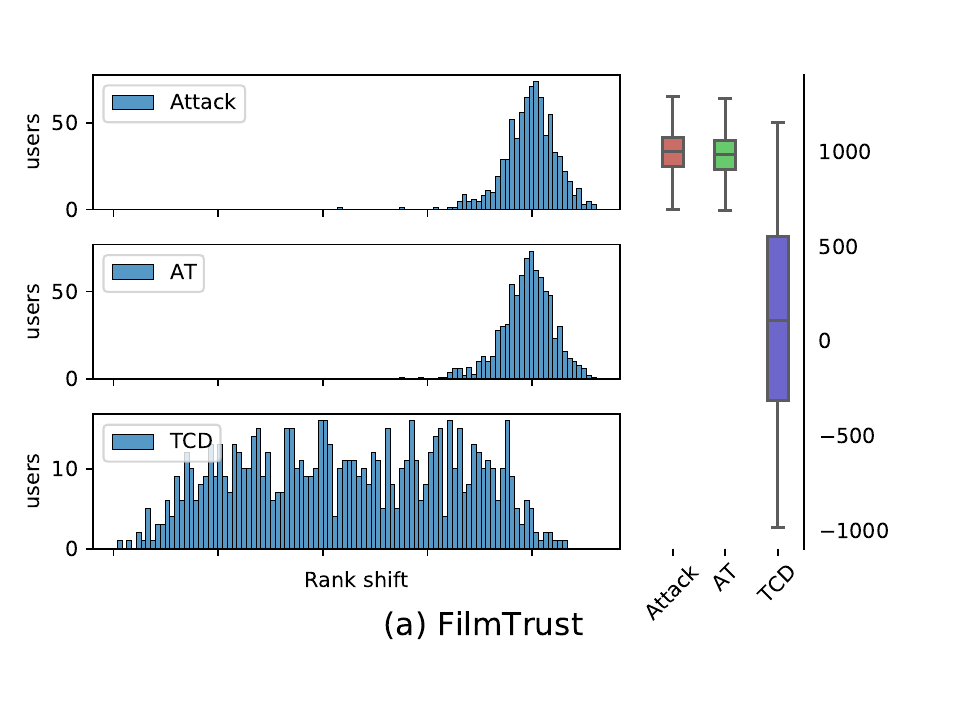}\quad}
{\includegraphics[width=0.48\textwidth]{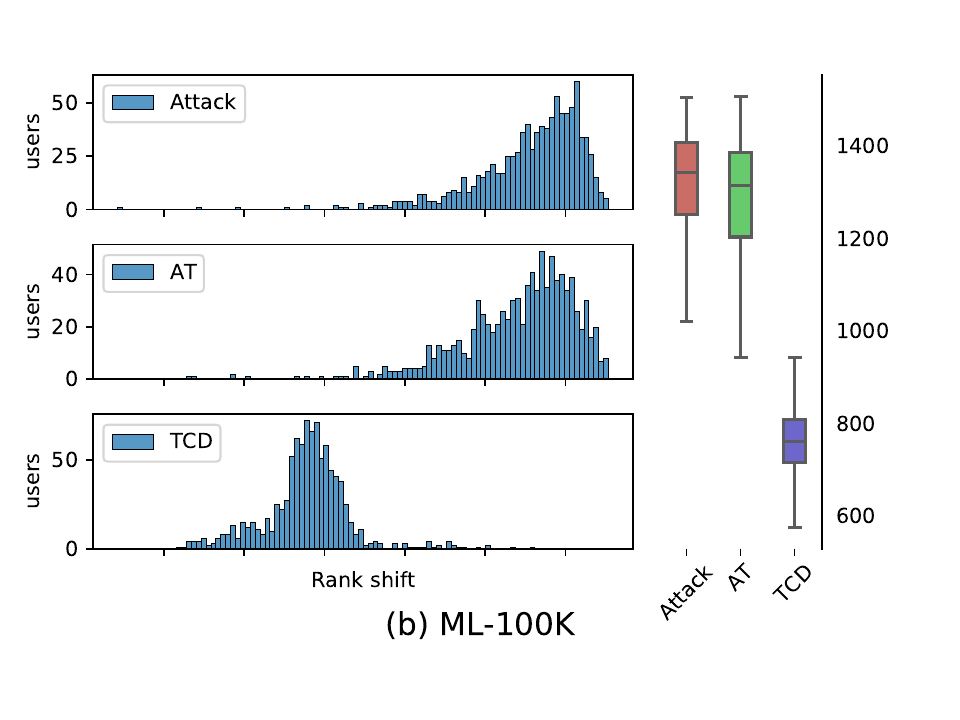}\par\medskip}
{\includegraphics[width=0.48\textwidth]{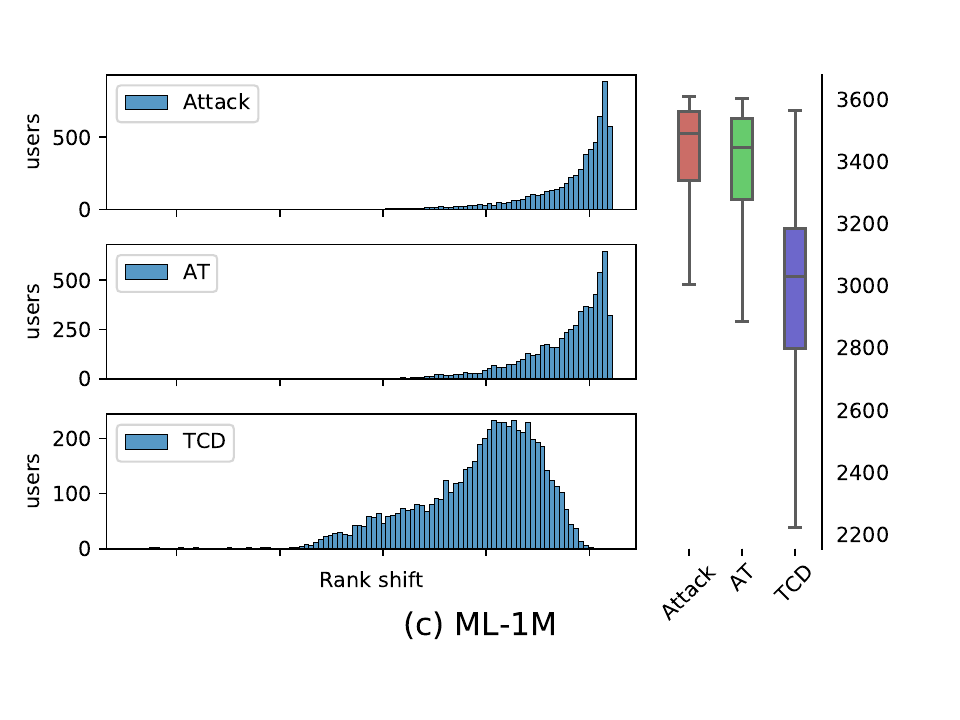}}
\caption{Rank shift distribution of target items (unpopular items). The smaller the rank shift, the smaller the impact of the attack.}
\label{fig: defense_distribution}
\end{figure*}

Besides, Fig. \ref{fig: defense_distribution} shows the Rank shift distribution of target items (unpopular items) under the TNA attack. The attack significantly promotes the target item’s
rank among all users. After applying adversarial training, the rank change caused by the attack is mitigated but remains slight. On the contrary, TCD clearly drives the distribution of rank shift toward 0, indicating that TCD can produce more stable recommendations in a perturbed environment.
In conclusion, these results provide strong evidence of TCD's positive impact on enhancing recommendation system robustness against poisoning attacks.

\subsubsection{Generalization}
\begin{table*}[htbp]
  \centering
  \small
  \caption{The performance in the test set (generalization). *, **, and *** indicate that the improvements over the unperturbed model are statistically significant for $ p<0.05, p<0.01 $, and $ p<0.001 $, respectively.}
  \renewcommand\arraystretch{0.85}
  \begin{adjustbox}{width=1\textwidth}
    \begin{tabular}{c|c|cccccccc}
    \hline \noalign{\smallskip}
    \multirow{2}[0]{*}{Dataset} & \multirow{2}[0]{*}{Method} & \multicolumn{8}{c}{Random Items} \\
    \cmidrule{3-10}
          &       & Average & Random & AUSH  & PGA   & TNA   & DL    & CoAttack & GCoAttack \\
    \noalign{\smallskip} \hline \noalign{\smallskip}
    \multirow{6}[0]{*}{FilmTrust}  & Origin& 0.8485 & 0.8470 & 0.8434 & 0.8533 & 0.8476 & 0.8476 & 0.8464 & 0.8472  \\
    & Attack & 0.8366 & 0.8351 & 0.8383 & 0.8360 & 0.8251 & 0.8373 & 0.8347 & 0.83251 \\
          & AT    & 0.8130 & 0.8076 & 0.8152 & 0.8083 & 0.7897 & 0.8109 & 0.8003 & 0.7964 \\
          & RAT  & 0.8314 & 0.8351 & 0.8357 & 0.8335 & 0.8214 & 0.8322 & 0.8330 & 0.8299 \\
          & TCD    & \textbf{0.8719} & \textbf{0.8720} & \textbf{0.8728} & \textbf{0.8725} & \textbf{0.8695} & \textbf{0.8724} & \textbf{0.8724} & \textbf{0.8714} \\
          \cmidrule{2-10}
          & p-value & ***   & ***   & ***   & ***   & ***   & ***   & ***   & *** \\
          \midrule
    \multirow{6}[0]{*}{ML-100K} & Origin & 0.2364 & 0.2478 & 0.2363 & 0.2387 & 0.2356 & 0.2334 & 0.2362 & 0.2337  \\
    & Attack & 0.2259 & 0.2368 & 0.2257 & 0.2235 & 0.2240 & 0.2276 & 0.2287 & 0.2284\\
          & AT   & 0.2165 & 0.2254 & 0.2124 & 0.2083 & 0.2121 & 0.2204 & 0.2196 & 0.2173 \\
          & RAT   & 0.2226 & 0.2314 & 0.2218 & 0.2193 & 0.2216 & 0.2248 & 0.2261 & 0.2258 \\
          & TCD   & \textbf{0.3179} & \textbf{0.3222} & \textbf{0.3165} & \textbf{0.3231} & \textbf{0.3223} & \textbf{0.3131} & \textbf{0.3140} & \textbf{0.3083}\\
          \cmidrule{2-10}
          & p-value & ***   & ***   & ***   & ***   & ***   & ***   & ***   & *** \\
          \midrule
    \multirow{6}[0]{*}{ML-1M} & Origin & 0.1034 & 0.1044 & 0.1037 & 0.1037 & 0.1027 & 0.1035 & 0.1032 & 0.1028\\
          & Attack & 0.0859 & 0.1007 & 0.0848 & 0.0961 & 0.0868 & 0.0988 & 0.0974 & 0.0991\\
          & AT    & 0.0494 & 0.0978 & 0.0458 & 0.0909 & 0.0470 & 0.1001 & 0.0966 & 0.0969 \\
          & RAT   & 0.0825 & 0.0997 & 0.0811 & 0.0940 & 0.0818 & 0.0987 & 0.0960 & 0.0965  \\
          & TCD  &  \textbf{0.1256} & \textbf{0.1276} & \textbf{0.1264} & \textbf{0.1255} & \textbf{0.1262} & \textbf{0.1267} & \textbf{0.1267} & \textbf{0.1238}\\
          \cmidrule{2-10}
          & p-value & ***   & ***   & ***   & ***   & ***   & ***   & ***   & *** \\
          \midrule
          \midrule
    \multirow{2}[0]{*}{Dataset} & \multirow{2}[0]{*}{Method} & \multicolumn{8}{c}{Unpopular Items} \\
    \cmidrule{3-10}
          &       & Average & Random & AUSH  & PGA   & TNA   & DL    & COATK & GCOATK \\
    \midrule
    \multirow{6}[0]{*}{FilmTrust} & Origin & 0.8505 & 0.8430 & 0.8417 & 0.8518 & 0.8467 & 0.8467 & 0.8367 & 0.8464\\
          & Attack  & 0.8374 & 0.8340 & 0.8350 & 0.8394 & 0.8258 & 0.8373 & 0.8367 & 0.8329 \\
          & AT  & 0.8039 & 0.8056 & 0.8046 & 0.8077 & 0.7855 & 0.8050 & 0.8099 & 0.8012  \\
          & RAT    & 0.8341 & 0.8327 & 0.8327 & 0.8348 & 0.8180 & 0.8349 & 0.8334 & 0.8299\\
          & TCD  &  \textbf{0.8724} & \textbf{0.8725} & \textbf{0.8725} & \textbf{0.8721} & \textbf{0.8714} & \textbf{0.8715} & \textbf{0.8718} & \textbf{0.8722} \\
          \cmidrule{2-10}
          & p-value & ***   & ***   & ***   & ***   & ***   & ***   & ***   & *** \\
          \midrule
    \multirow{6}[0]{*}{ML-100K} & Origin  & 0.2375 & 0.2598 & 0.2354 & 0.2390 & 0.2314 & 0.2311 & 0.2310 & 0.2365  \\
    & Attack & 0.2248 & 0.2317 & 0.2258 & 0.2154 & 0.2247 & 0.2205 & 0.2198 & 0.2163\\
          & AT     & 0.2171 & 0.2232 & 0.2174 & 0.2075 & 0.2094 & 0.2129 & 0.2077 & 0.2088 \\
          & RAT  & 0.2269 & 0.2301 & 0.2270 & 0.2124 & 0.2213 & 0.2168 & 0.2187 & 0.2184\\
          & TCD   & \textbf{0.3117} & \textbf{0.3105} & \textbf{0.3112} & \textbf{0.3179} & \textbf{0.3134} & \textbf{0.3166} & \textbf{0.3286} & \textbf{0.3132} \\
          \cmidrule{2-10}
          & p-value & ***   & ***   & ***   & ***   & ***   & ***   & ***   & *** \\
          \midrule
    \multirow{6}[0]{*}{ML-1M} & Origin & 0.1026 & 0.1033 & 0.1032 & 0.1031 & 0.1046 & 0.1036 & 0.1036 & 0.1038 \\
          & Attack & 0.0836 & 0.0965 & 0.0808 & 0.0902 & 0.0870 & 0.0949 & 0.0926 & 0.0938\\
          & AT   & 0.0441 & 0.0956 & 0.0360 & 0.0820 & 0.0450 & 0.0938 & 0.0900 & 0.0910 \\
          & RAT  & 0.0809 & 0.0954 & 0.0743 & 0.0886 & 0.0827 & 0.0935 & 0.0915 & 0.0916 \\
          & TCD   & \textbf{0.1263} & \textbf{0.1250} & \textbf{0.1265} & \textbf{0.1260} & \textbf{0.1269} & \textbf{0.1261} & \textbf{0.1210} & \textbf{0.1199}\\
          \cmidrule{2-10}
          & p-value & ***   & ***   & ***   & ***   & ***   & ***   & ***   & *** \\
          \midrule
    \end{tabular}%
    \end{adjustbox}
  \label{tab:defense_generalization}%
\end{table*}%
A desirable defense should enhance robustness while preserving the model's generalization performance. Robustness achieved at the expense of standard generalization is meaningless. Therefore, in this section, we evaluate the generalization of the recommendation system (i.e., performance on the test set) under various defense strategies, as shown in Table \ref{fig: defense_distribution}. On the one hand, it can be seen that adding adversarial noise to the model parameters through Adversarial Training (AT) reduces the model's performance, which is consistent with existing findings that adversarial training cannot simultaneously enjoy both robustness and generalization \cite{raghunathan2019adversarial}. On the other hand, the proposed TCD does not compromise the model's robustness and even improves its recommendation performance. For instance, on ML-100K, it elevates the HR@50 from 0.2364 to 0.32, demonstrating the advantage of collaborative training among the three models.

\subsubsection{Performance under Different Attack knowledge-cost}
\begin{figure}[h]
	\centering
	\includegraphics[width=1.\columnwidth]{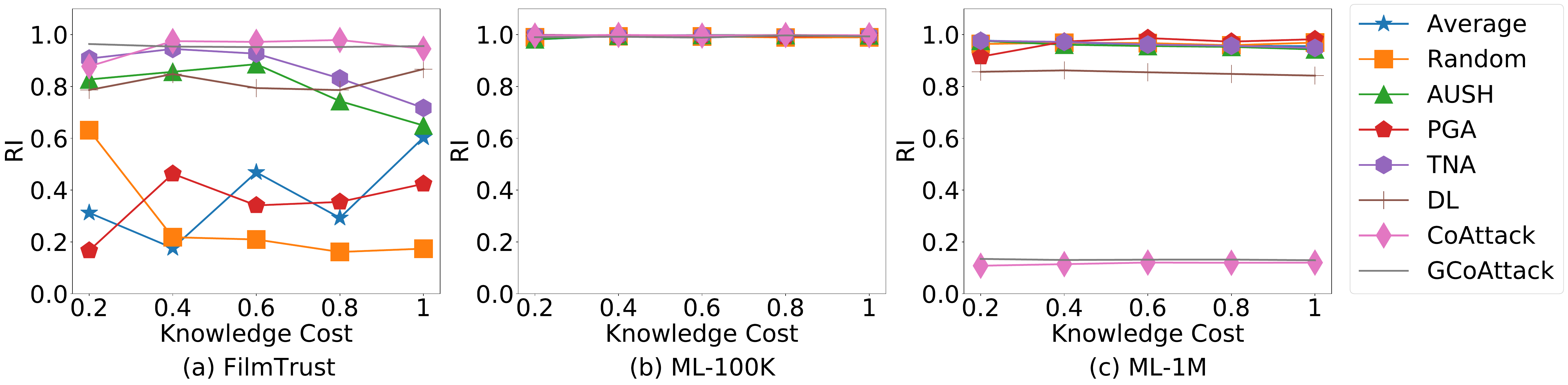}
	\caption{Robustness improvement under different attack knowledge-cost.
	}
	\label{fig:knowledge cost}
\end{figure}

As verified in Section \ref{sec: attack_comparision}, as the knowledge available to the attacker increases, the damage to the model will also be greater. In this section, we explore the robustness improvement of TCD under different knowledge, as shown in Fig. \ref{fig:knowledge cost}. First, the impact of the attacker's knowledge on the defensive performance is minimal, indicating that our attack possesses universality, even if we are unaware of the specific configuration employed by the attacker. Second, TCD consistently delivers satisfactory results against most attacks. This underscores the potential applicability of our algorithm in real-world systems, where defenders often lack precise knowledge of the attack algorithms used by adversaries.

\subsubsection{Performance under Different Pseudo-label Ratios}
\label{sec: label ratio}
\begin{figure}[h]
	\centering
	\includegraphics[width=1\columnwidth]{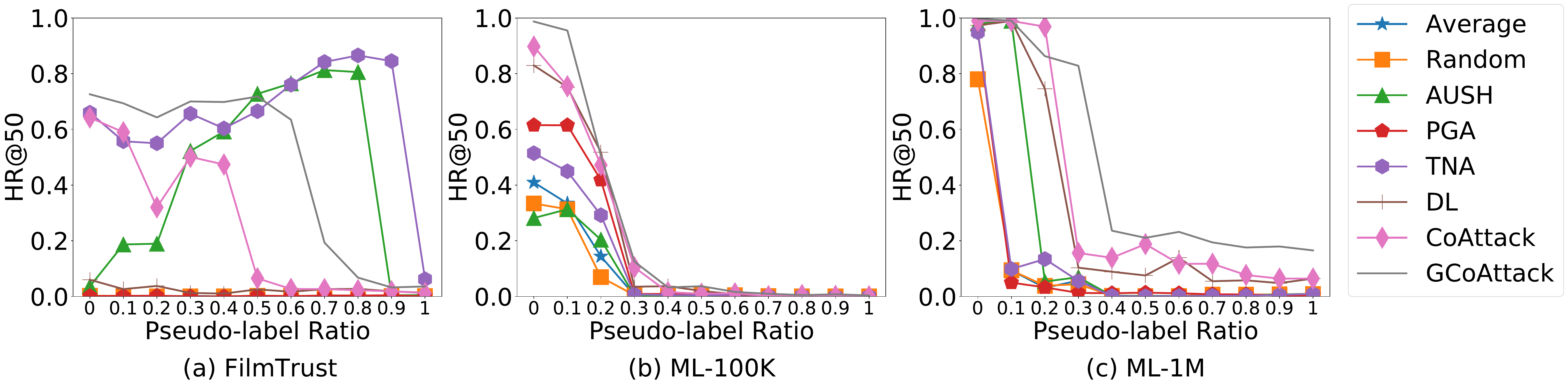}
	\caption{The defense performance (unpopular items) on ML-1M under different injected pseudo-label ratios.}
	\label{fig:pseudo labels rate}
\end{figure}

The training time of TCD is directly proportional to the size of the training dataset, which means proportional to the number of injected pseudo-labels. Therefore, under the default settings of the experiment, we tolerate all high-confidence pseudo-labels for the smaller dataset (FilmTrust and ML-100K), while for the larger ML-1M, we only use 20\% of the data to improve the training efficiency. In this section, we analyze the impact of different pseudo-label injection ratios on the model robustness, as shown in Fig. \ref{fig:pseudo labels rate}. Overall, the model robustness increases as the number of injected pseudo-labels increases. It is worth noting that in the larger ML-1M dataset, we observe that when the injection ratio is between 20\% and 30\%, the model's robustness against attacks has already reached a satisfactory level. This desirable property makes applying TCD to large-scale datasets in practice feasible. Additionally, we find that in FilmTrust, the model's robustness against AUSH decreases with fewer pseudo-labels. We suspect that AUSH aims to generate profiles that are confusingly similar to real profiles, which may cause the high-confidence pseudo-labels to be mixed with false ones; that is, it may inject more fraudulent data and lead to a decline in robustness. As the number of pseudo-labels increases, the number of trusted labels also grows, gradually diminishing the impact of the fake data. This finding underscores the importance of studying imperceptible attacks, which will be a focus for our future work.

\section{Conclusion and Outlook}\label{sec7}
In this paper, we first proposed a novel defense method, Triple Cooperative Defense (TCD), to enhance recommendation robustness against poisoning attacks. TCD integrates data processing and model robustness boosting by using three recommendation models for cooperative training. The high-confidence prediction ratings of any two models are used as auxiliary training data for the remaining model in each round of training. Second, we revisited the poisoning attack and proposed an efficient poisoning attack, Co-training Attack (CoAttack), which cooperatively optimizes attack objective and model training to generate malicious poisoning profiles efficiently. Additionally, we revealed that existing attacks are usually optimized based on an optimistic, defenseless model, which limits the attack performance. To this end, we further proposed a more harmful attack, a Game-based Co-training Attack (GCoAttack), to train the proposed TCD and CoAttack cooperatively. Extensive experiments on three datasets demonstrate the effectiveness of the proposed methods over state-of-the-art baselines.

In the future, we may conduct further work from two aspects: (1) Defense Enhancement: The Triple Cooperative Defense (TCD) method introduced in this paper presents a versatile framework that can be integrated with other defense strategies. We plan to explore the application of TCD beyond recommendation systems, potentially extending its use to other domains and fields where robustness against adversarial attacks is critical. (2) Advancements in Attack-Defense Dynamics: In our exploration of Game-based Co-training Attack (GCoAttack), we demonstrated that an attack-defense game can maximize the threat of an attack. In the future, we aim to investigate whether such dynamic interactions can also lead to enhanced defense capabilities. This research will delve deeper into understanding the intricate balance between attack and defense in adversarial settings.

\section*{Acknowledgment}

The work was supported by grants from the National Natural Science Foundation of China (No. 62022077).

\section{Declarations}
\subsection{Ethical Approval}
Not applicable. I declare that this paper does not involve any human or animal studies, so no ethical issues are involved.
\subsection{Competing interests}
I declare that the authors have no competing interests as defined by Springer, or other interests that might be perceived to influence the results and/or discussion reported in this paper.
\subsection{Authors' contributions}
Qingyang Wang and Chenwang Wu contribute equally to this paper, including algorithm implementation, experimental data collation, and paper writing. Defu Lian and Enhong Chen proofread the manuscript. In addition, all authors reviewed the manuscript.
\subsection{Funding}
The work was supported by grants from the National Natural Science Foundation of China (No. 62022077).
\subsection{Availability of data and materials}
The source code and data are available at \url{https://github.com/greensun0830/Cotraining-Attack}.

\bibliography{sn-bibliography}


\begin{thebibliography}{58}
\ifx \bisbn   \undefined \def \bisbn  #1{ISBN #1}\fi
\ifx \binits  \undefined \def \binits#1{#1}\fi
\ifx \bauthor  \undefined \def \bauthor#1{#1}\fi
\ifx \batitle  \undefined \def \batitle#1{#1}\fi
\ifx \bjtitle  \undefined \def \bjtitle#1{#1}\fi
\ifx \bvolume  \undefined \def \bvolume#1{\textbf{#1}}\fi
\ifx \byear  \undefined \def \byear#1{#1}\fi
\ifx \bissue  \undefined \def \bissue#1{#1}\fi
\ifx \bfpage  \undefined \def \bfpage#1{#1}\fi
\ifx \blpage  \undefined \def \blpage #1{#1}\fi
\ifx \burl  \undefined \def \burl#1{\textsf{#1}}\fi
\ifx \doiurl  \undefined \def \doiurl#1{\url{https://doi.org/#1}}\fi
\ifx \betal  \undefined \def \betal{\textit{et al.}}\fi
\ifx \binstitute  \undefined \def \binstitute#1{#1}\fi
\ifx \binstitutionaled  \undefined \def \binstitutionaled#1{#1}\fi
\ifx \bctitle  \undefined \def \bctitle#1{#1}\fi
\ifx \beditor  \undefined \def \beditor#1{#1}\fi
\ifx \bpublisher  \undefined \def \bpublisher#1{#1}\fi
\ifx \bbtitle  \undefined \def \bbtitle#1{#1}\fi
\ifx \bedition  \undefined \def \bedition#1{#1}\fi
\ifx \bseriesno  \undefined \def \bseriesno#1{#1}\fi
\ifx \blocation  \undefined \def \blocation#1{#1}\fi
\ifx \bsertitle  \undefined \def \bsertitle#1{#1}\fi
\ifx \bsnm \undefined \def \bsnm#1{#1}\fi
\ifx \bsuffix \undefined \def \bsuffix#1{#1}\fi
\ifx \bparticle \undefined \def \bparticle#1{#1}\fi
\ifx \barticle \undefined \def \barticle#1{#1}\fi
\bibcommenthead
\ifx \bconfdate \undefined \def \bconfdate #1{#1}\fi
\ifx \botherref \undefined \def \botherref #1{#1}\fi
\ifx \url \undefined \def \url#1{\textsf{#1}}\fi
\ifx \bchapter \undefined \def \bchapter#1{#1}\fi
\ifx \bbook \undefined \def \bbook#1{#1}\fi
\ifx \bcomment \undefined \def \bcomment#1{#1}\fi
\ifx \oauthor \undefined \def \oauthor#1{#1}\fi
\ifx \citeauthoryear \undefined \def \citeauthoryear#1{#1}\fi
\ifx \endbibitem  \undefined \def \endbibitem {}\fi
\ifx \bconflocation  \undefined \def \bconflocation#1{#1}\fi
\ifx \arxivurl  \undefined \def \arxivurl#1{\textsf{#1}}\fi
\csname PreBibitemsHook\endcsname

\bibitem[\protect\citeauthoryear{Bobadilla
  et~al.}{2013}]{bobadilla2013recommender}
\begin{barticle}
\bauthor{\bsnm{Bobadilla}, \binits{J.}},
\bauthor{\bsnm{Ortega}, \binits{F.}},
\bauthor{\bsnm{Hernando}, \binits{A.}},
\bauthor{\bsnm{Guti{\'e}rrez}, \binits{A.}}:
\batitle{Recommender systems survey}.
\bjtitle{Knowledge-based systems}
\bvolume{46},
\bfpage{109}--\blpage{132}
(\byear{2013})
\end{barticle}
\endbibitem

\bibitem[\protect\citeauthoryear{Himeur et~al.}{2022}]{himeur2022blockchain}
\begin{barticle}
\bauthor{\bsnm{Himeur}, \binits{Y.}},
\bauthor{\bsnm{Sayed}, \binits{A.}},
\bauthor{\bsnm{Alsalemi}, \binits{A.}},
\bauthor{\bsnm{Bensaali}, \binits{F.}},
\bauthor{\bsnm{Amira}, \binits{A.}},
\bauthor{\bsnm{Varlamis}, \binits{I.}},
\bauthor{\bsnm{Eirinaki}, \binits{M.}},
\bauthor{\bsnm{Sardianos}, \binits{C.}},
\bauthor{\bsnm{Dimitrakopoulos}, \binits{G.}}:
\batitle{Blockchain-based recommender systems: Applications, challenges and
  future opportunities}.
\bjtitle{Computer Science Review}
\bvolume{43},
\bfpage{100439}
(\byear{2022})
\end{barticle}
\endbibitem

\bibitem[\protect\citeauthoryear{Lian et~al.}{2020}]{lian2020geography}
\begin{bchapter}
\bauthor{\bsnm{Lian}, \binits{D.}},
\bauthor{\bsnm{Wu}, \binits{Y.}},
\bauthor{\bsnm{Ge}, \binits{Y.}},
\bauthor{\bsnm{Xie}, \binits{X.}},
\bauthor{\bsnm{Chen}, \binits{E.}}:
\bctitle{Geography-aware sequential location recommendation}.
In: \bbtitle{Proceedings of KDD'20},
pp. \bfpage{2009}--\blpage{2019}
(\byear{2020})
\end{bchapter}
\endbibitem

\bibitem[\protect\citeauthoryear{Chevalier and
  Mayzlin}{2006}]{chevalier2006effect}
\begin{barticle}
\bauthor{\bsnm{Chevalier}, \binits{J.A.}},
\bauthor{\bsnm{Mayzlin}, \binits{D.}}:
\batitle{The effect of word of mouth on sales: Online book reviews}.
\bjtitle{J Mark Res}
\bvolume{43}(\bissue{3}),
\bfpage{345}--\blpage{354}
(\byear{2006})
\end{barticle}
\endbibitem

\bibitem[\protect\citeauthoryear{Wu et~al.}{2021}]{wu2021triple}
\begin{bchapter}
\bauthor{\bsnm{Wu}, \binits{C.}},
\bauthor{\bsnm{Lian}, \binits{D.}},
\bauthor{\bsnm{Ge}, \binits{Y.}},
\bauthor{\bsnm{Zhu}, \binits{Z.}},
\bauthor{\bsnm{Chen}, \binits{E.}}:
\bctitle{Triple adversarial learning for influence based poisoning attack in
  recommender systems}.
In: \bbtitle{Proceedings of KDD'21},
pp. \bfpage{1830}--\blpage{1840}
(\byear{2021})
\end{bchapter}
\endbibitem

\bibitem[\protect\citeauthoryear{Li et~al.}{2016}]{li2016data}
\begin{barticle}
\bauthor{\bsnm{Li}, \binits{B.}},
\bauthor{\bsnm{Wang}, \binits{Y.}},
\bauthor{\bsnm{Singh}, \binits{A.}},
\bauthor{\bsnm{Vorobeychik}, \binits{Y.}}:
\batitle{Data poisoning attacks on factorization-based collaborative
  filtering}.
\bjtitle{NIPS}
\bvolume{29},
\bfpage{1885}--\blpage{1893}
(\byear{2016})
\end{barticle}
\endbibitem

\bibitem[\protect\citeauthoryear{Lin et~al.}{2020}]{lin2020attacking}
\begin{bchapter}
\bauthor{\bsnm{Lin}, \binits{C.}},
\bauthor{\bsnm{Chen}, \binits{S.}},
\bauthor{\bsnm{Li}, \binits{H.}},
\bauthor{\bsnm{Xiao}, \binits{Y.}},
\bauthor{\bsnm{Li}, \binits{Q.} \bsuffix{Lianyun \textbf{}and~Yang}}:
\bctitle{Attacking recommender systems with augmented user profiles}.
In: \bbtitle{CIKM},
pp. \bfpage{855}--\blpage{864}
(\byear{2020})
\end{bchapter}
\endbibitem

\bibitem[\protect\citeauthoryear{Liu et~al.}{2014}]{liu2014new}
\begin{barticle}
\bauthor{\bsnm{Liu}, \binits{H.}},
\bauthor{\bsnm{Hu}, \binits{Z.}},
\bauthor{\bsnm{Mian}, \binits{A.}},
\bauthor{\bsnm{Tian}, \binits{H.}},
\bauthor{\bsnm{Zhu}, \binits{X.}}:
\batitle{A new user similarity model to improve the accuracy of collaborative
  filtering}.
\bjtitle{KBS}
\bvolume{56},
\bfpage{156}--\blpage{166}
(\byear{2014})
\end{barticle}
\endbibitem

\bibitem[\protect\citeauthoryear{Madry et~al.}{2017}]{madry2017towards}
\begin{botherref}
\oauthor{\bsnm{Madry}, \binits{A.}},
\oauthor{\bsnm{Makelov}, \binits{A.}},
\oauthor{\bsnm{Schmidt}, \binits{L.}},
\oauthor{\bsnm{Tsipras}, \binits{D.}},
\oauthor{\bsnm{Vladu}, \binits{A.}}:
Towards deep learning models resistant to adversarial attacks.
arXiv
(2017)
\end{botherref}
\endbibitem

\bibitem[\protect\citeauthoryear{Wu et~al.}{2021}]{wu2021fight}
\begin{bchapter}
\bauthor{\bsnm{Wu}, \binits{C.}},
\bauthor{\bsnm{Lian}, \binits{D.}},
\bauthor{\bsnm{Ge}, \binits{Y.}},
\bauthor{\bsnm{Zhu}, \binits{Z.}},
\bauthor{\bsnm{Chen}, \binits{E.}},
\bauthor{\bsnm{Yuan}, \binits{S.}}:
\bctitle{Fight fire with fire: Towards robust recommender systems via
  adversarial poisoning training}.
In: \bbtitle{SIGIR},
pp. \bfpage{1074}--\blpage{1083}
(\byear{2021})
\end{bchapter}
\endbibitem

\bibitem[\protect\citeauthoryear{Nguyen~Thanh
  et~al.}{2023}]{nguyen2023poisoning}
\begin{barticle}
\bauthor{\bsnm{Nguyen~Thanh}, \binits{T.}},
\bauthor{\bsnm{Quach}, \binits{N.D.K.}},
\bauthor{\bsnm{Nguyen}, \binits{T.T.}},
\bauthor{\bsnm{Huynh}, \binits{T.T.}},
\bauthor{\bsnm{Vu}, \binits{V.H.}},
\bauthor{\bsnm{Nguyen}, \binits{P.L.}},
\bauthor{\bsnm{Jo}, \binits{J.}},
\bauthor{\bsnm{Nguyen}, \binits{Q.V.H.}}:
\batitle{Poisoning gnn-based recommender systems with generative
  surrogate-based attacks}.
\bjtitle{ACM Transactions on Information Systems}
\bvolume{41}(\bissue{3}),
\bfpage{1}--\blpage{24}
(\byear{2023})
\end{barticle}
\endbibitem

\bibitem[\protect\citeauthoryear{Lam and Riedl}{2004}]{lam2004shilling}
\begin{bchapter}
\bauthor{\bsnm{Lam}, \binits{S.K.}},
\bauthor{\bsnm{Riedl}, \binits{J.}}:
\bctitle{Shilling recommender systems for fun and profit}.
In: \bbtitle{WWW},
pp. \bfpage{393}--\blpage{402}
(\byear{2004})
\end{bchapter}
\endbibitem

\bibitem[\protect\citeauthoryear{Burke et~al.}{2006}]{burke2006classification}
\begin{bchapter}
\bauthor{\bsnm{Burke}, \binits{R.}},
\bauthor{\bsnm{Mobasher}, \binits{B.}},
\bauthor{\bsnm{Williams}, \binits{C.}},
\bauthor{\bsnm{Bhaumik}, \binits{R.}}:
\bctitle{Classification features for attack detection in collaborative
  recommender systems}.
In: \bbtitle{KDD},
pp. \bfpage{542}--\blpage{547}
(\byear{2006})
\end{bchapter}
\endbibitem

\bibitem[\protect\citeauthoryear{Cohen et~al.}{2021}]{cohen2021black}
\begin{bchapter}
\bauthor{\bsnm{Cohen}, \binits{R.}},
\bauthor{\bsnm{Sar~Shalom}, \binits{O.}},
\bauthor{\bsnm{Jannach}, \binits{D.}},
\bauthor{\bsnm{Amir}, \binits{A.}}:
\bctitle{A black-box attack model for visually-aware recommender systems}.
In: \bbtitle{Proceedings of the 14th ACM International Conference on Web Search
  and Data Mining},
pp. \bfpage{94}--\blpage{102}
(\byear{2021})
\end{bchapter}
\endbibitem

\bibitem[\protect\citeauthoryear{Yue et~al.}{2021}]{yue2021black}
\begin{bchapter}
\bauthor{\bsnm{Yue}, \binits{Z.}},
\bauthor{\bsnm{He}, \binits{Z.}},
\bauthor{\bsnm{Zeng}, \binits{H.}},
\bauthor{\bsnm{McAuley}, \binits{J.}}:
\bctitle{Black-box attacks on sequential recommenders via data-free model
  extraction}.
In: \bbtitle{Proceedings of the 15th ACM Conference on Recommender Systems},
pp. \bfpage{44}--\blpage{54}
(\byear{2021})
\end{bchapter}
\endbibitem

\bibitem[\protect\citeauthoryear{Zhang et~al.}{2022}]{zhang2022pipattack}
\begin{bchapter}
\bauthor{\bsnm{Zhang}, \binits{S.}},
\bauthor{\bsnm{Yin}, \binits{H.}},
\bauthor{\bsnm{Chen}, \binits{T.}},
\bauthor{\bsnm{Huang}, \binits{Z.}},
\bauthor{\bsnm{Nguyen}, \binits{Q.V.H.}},
\bauthor{\bsnm{Cui}, \binits{L.}}:
\bctitle{Pipattack: Poisoning federated recommender systems for manipulating
  item promotion}.
In: \bbtitle{Proceedings of the Fifteenth ACM International Conference on Web
  Search and Data Mining},
pp. \bfpage{1415}--\blpage{1423}
(\byear{2022})
\end{bchapter}
\endbibitem

\bibitem[\protect\citeauthoryear{Fang et~al.}{2018}]{fang2018poisoning}
\begin{bchapter}
\bauthor{\bsnm{Fang}, \binits{M.}},
\bauthor{\bsnm{Yang}, \binits{G.}},
\bauthor{\bsnm{Gong}, \binits{N.Z.}},
\bauthor{\bsnm{Liu}, \binits{J.}}:
\bctitle{Poisoning attacks to graph-based recommender systems}.
In: \bbtitle{Proceedings of the 34th Annual Computer Security Applications
  Conference},
pp. \bfpage{381}--\blpage{392}
(\byear{2018})
\end{bchapter}
\endbibitem

\bibitem[\protect\citeauthoryear{Huang et~al.}{2021}]{huang2021data}
\begin{botherref}
\oauthor{\bsnm{Huang}, \binits{H.}},
\oauthor{\bsnm{Mu}, \binits{J.}},
\oauthor{\bsnm{Gong}, \binits{N.Z.}},
\oauthor{\bsnm{Li}, \binits{Q.}},
\oauthor{\bsnm{Liu}, \binits{B.}},
\oauthor{\bsnm{Xu}, \binits{M.}}:
Data poisoning attacks to deep learning based recommender systems.
arXiv preprint arXiv:2101.02644
(2021)
\end{botherref}
\endbibitem

\bibitem[\protect\citeauthoryear{Wang et~al.}{2022}]{wang2022towards}
\begin{bchapter}
\bauthor{\bsnm{Wang}, \binits{Q.}},
\bauthor{\bsnm{Lian}, \binits{D.}},
\bauthor{\bsnm{Wu}, \binits{C.}},
\bauthor{\bsnm{Chen}, \binits{E.}}:
\bctitle{Towards robust recommender systems via triple cooperative defense}.
In: \bbtitle{Web Information Systems Engineering--WISE 2022: 23rd International
  Conference, Biarritz, France, November 1--3, 2022, Proceedings},
pp. \bfpage{564}--\blpage{578}
(\byear{2022}).
\bcomment{Springer}
\end{bchapter}
\endbibitem

\bibitem[\protect\citeauthoryear{Du et~al.}{2018}]{du2018enhancing}
\begin{barticle}
\bauthor{\bsnm{Du}, \binits{Y.}},
\bauthor{\bsnm{Fang}, \binits{M.}},
\bauthor{\bsnm{Yi}, \binits{J.}},
\bauthor{\bsnm{Xu}, \binits{C.}},
\bauthor{\bsnm{Cheng}, \binits{J.}},
\bauthor{\bsnm{Tao}, \binits{D.}}:
\batitle{Enhancing the robustness of neural collaborative filtering systems
  under malicious attacks}.
\bjtitle{IEEE Trans. Multimedia}
\bvolume{21}(\bissue{3}),
\bfpage{555}--\blpage{565}
(\byear{2018})
\end{barticle}
\endbibitem

\bibitem[\protect\citeauthoryear{Si and Li}{2020}]{si2020shilling}
\begin{barticle}
\bauthor{\bsnm{Si}, \binits{M.}},
\bauthor{\bsnm{Li}, \binits{Q.}}:
\batitle{Shilling attacks against collaborative recommender systems: a review}.
\bjtitle{Artif Intell Rev}
\bvolume{53}(\bissue{1}),
\bfpage{291}--\blpage{319}
(\byear{2020})
\end{barticle}
\endbibitem

\bibitem[\protect\citeauthoryear{Ovaisi et~al.}{2022}]{ovaisi2022rgrecsys}
\begin{botherref}
\oauthor{\bsnm{Ovaisi}, \binits{Z.}},
\oauthor{\bsnm{Heinecke}, \binits{S.}},
\oauthor{\bsnm{Li}, \binits{J.}},
\oauthor{\bsnm{Zhang}, \binits{Y.}},
\oauthor{\bsnm{Zheleva}, \binits{E.}},
\oauthor{\bsnm{Xiong}, \binits{C.}}:
Rgrecsys: A toolkit for robustness evaluation of recommender systems.
arXiv
(2022)
\end{botherref}
\endbibitem

\bibitem[\protect\citeauthoryear{Chen et~al.}{2021}]{chen2021data}
\begin{barticle}
\bauthor{\bsnm{Chen}, \binits{L.}},
\bauthor{\bsnm{Xu}, \binits{Y.}},
\bauthor{\bsnm{Xie}, \binits{F.}},
\bauthor{\bsnm{Huang}, \binits{M.}},
\bauthor{\bsnm{Zheng}, \binits{Z.}}:
\batitle{Data poisoning attacks on neighborhood-based recommender systems}.
\bjtitle{Transactions on Emerging Telecommunications Technologies}
\bvolume{32}(\bissue{6}),
\bfpage{3872}
(\byear{2021})
\end{barticle}
\endbibitem

\bibitem[\protect\citeauthoryear{Guo et~al.}{2017}]{guo2017deepfm}
\begin{botherref}
\oauthor{\bsnm{Guo}, \binits{H.}},
\oauthor{\bsnm{Tang}, \binits{R.}},
\oauthor{\bsnm{Ye}, \binits{Y.}},
\oauthor{\bsnm{Li}, \binits{Z.}},
\oauthor{\bsnm{He}, \binits{X.}}:
Deepfm: a factorization-machine based neural network for ctr prediction.
arXiv
(2017)
\end{botherref}
\endbibitem

\bibitem[\protect\citeauthoryear{He et~al.}{2017}]{he2017neural}
\begin{bchapter}
\bauthor{\bsnm{He}, \binits{X.}},
\bauthor{\bsnm{Liao}, \binits{L.}},
\bauthor{\bsnm{Zhang}, \binits{H.}},
\bauthor{\bsnm{Nie}, \binits{L.}},
\bauthor{\bsnm{Hu}, \binits{X.}},
\bauthor{\bsnm{Chua}, \binits{T.-S.}}:
\bctitle{Neural collaborative filtering}.
In: \bbtitle{WWW},
pp. \bfpage{173}--\blpage{182}
(\byear{2017})
\end{bchapter}
\endbibitem

\bibitem[\protect\citeauthoryear{Fang et~al.}{2020}]{fang2020influence}
\begin{bchapter}
\bauthor{\bsnm{Fang}, \binits{M.}},
\bauthor{\bsnm{Gong}, \binits{N.Z.}},
\bauthor{\bsnm{Liu}, \binits{J.}}:
\bctitle{Influence function based data poisoning attacks to top-n recommender
  systems}.
In: \bbtitle{Proceedings of The Web Conference 2020},
pp. \bfpage{3019}--\blpage{3025}
(\byear{2020})
\end{bchapter}
\endbibitem

\bibitem[\protect\citeauthoryear{Tang et~al.}{2020}]{tang2020revisiting}
\begin{bchapter}
\bauthor{\bsnm{Tang}, \binits{J.}},
\bauthor{\bsnm{Wen}, \binits{H.}},
\bauthor{\bsnm{Wang}, \binits{K.}}:
\bctitle{Revisiting adversarially learned injection attacks against recommender
  systems}.
In: \bbtitle{RecSys},
pp. \bfpage{318}--\blpage{327}
(\byear{2020})
\end{bchapter}
\endbibitem

\bibitem[\protect\citeauthoryear{Jin et~al.}{2020}]{jin2020sampling}
\begin{barticle}
\bauthor{\bsnm{Jin}, \binits{B.}},
\bauthor{\bsnm{Lian}, \binits{D.}},
\bauthor{\bsnm{Liu}, \binits{Z.}},
\bauthor{\bsnm{Liu}, \binits{Q.}},
\bauthor{\bsnm{Ma}, \binits{J.}},
\bauthor{\bsnm{Xie}, \binits{X.}},
\bauthor{\bsnm{Chen}, \binits{E.}}:
\batitle{Sampling-decomposable generative adversarial recommender}.
\bjtitle{Adv Neur In}
\bvolume{33},
\bfpage{22629}--\blpage{22639}
(\byear{2020})
\end{barticle}
\endbibitem

\bibitem[\protect\citeauthoryear{Christakopoulou and
  Banerjee}{2019}]{christakopoulou2019adversarial}
\begin{bchapter}
\bauthor{\bsnm{Christakopoulou}, \binits{K.}},
\bauthor{\bsnm{Banerjee}, \binits{A.}}:
\bctitle{Adversarial attacks on an oblivious recommender}.
In: \bbtitle{RecSys},
pp. \bfpage{322}--\blpage{330}
(\byear{2019})
\end{bchapter}
\endbibitem

\bibitem[\protect\citeauthoryear{Li et~al.}{2017}]{li2017triple}
\begin{botherref}
\oauthor{\bsnm{Li}, \binits{C.}},
\oauthor{\bsnm{Xu}, \binits{T.}},
\oauthor{\bsnm{Zhu}, \binits{J.}},
\oauthor{\bsnm{Zhang}, \binits{B.}}:
Triple generative adversarial nets.
Advances in neural information processing systems
\textbf{30}
(2017)
\end{botherref}
\endbibitem

\bibitem[\protect\citeauthoryear{Yang et~al.}{2017}]{yang2017fake}
\begin{bchapter}
\bauthor{\bsnm{Yang}, \binits{G.}},
\bauthor{\bsnm{Gong}, \binits{N.Z.}},
\bauthor{\bsnm{Cai}, \binits{Y.}}:
\bctitle{Fake co-visitation injection attacks to recommender systems.}
In: \bbtitle{NDSS}
(\byear{2017})
\end{bchapter}
\endbibitem

\bibitem[\protect\citeauthoryear{Oh and Kumar}{2022}]{oh2022robustness}
\begin{botherref}
\oauthor{\bsnm{Oh}, \binits{S.}},
\oauthor{\bsnm{Kumar}, \binits{S.}}:
Robustness of deep recommendation systems to untargeted interaction
  perturbations.
arXiv
(2022)
\end{botherref}
\endbibitem

\bibitem[\protect\citeauthoryear{Fan et~al.}{2021}]{fan2021attacking}
\begin{bchapter}
\bauthor{\bsnm{Fan}, \binits{W.}},
\bauthor{\bsnm{Derr}, \binits{T.}},
\bauthor{\bsnm{Zhao}, \binits{X.}},
\bauthor{\bsnm{Ma}, \binits{Y.}},
\bauthor{\bsnm{Liu}, \binits{H.}},
\bauthor{\bsnm{Wang}, \binits{J.}},
\bauthor{\bsnm{Tang}, \binits{J.}},
\bauthor{\bsnm{Li}, \binits{Q.}}:
\bctitle{Attacking black-box recommendations via copying cross-domain user
  profiles}.
In: \bbtitle{ICDE},
pp. \bfpage{1583}--\blpage{1594}
(\byear{2021}).
\bcomment{IEEE}
\end{bchapter}
\endbibitem

\bibitem[\protect\citeauthoryear{Song et~al.}{2020}]{song2020poisonrec}
\begin{bchapter}
\bauthor{\bsnm{Song}, \binits{J.}},
\bauthor{\bsnm{Li}, \binits{Z.}},
\bauthor{\bsnm{Hu}, \binits{Z.}},
\bauthor{\bsnm{Wu}, \binits{Y.}},
\bauthor{\bsnm{Li}, \binits{Z.}},
\bauthor{\bsnm{Li}, \binits{J.}},
\bauthor{\bsnm{Gao}, \binits{J.}}:
\bctitle{Poisonrec: an adaptive data poisoning framework for attacking
  black-box recommender systems}.
In: \bbtitle{ICDE},
pp. \bfpage{157}--\blpage{168}
(\byear{2020}).
\bcomment{IEEE}
\end{bchapter}
\endbibitem

\bibitem[\protect\citeauthoryear{Deldjoo et~al.}{2021}]{deldjoo2021survey}
\begin{barticle}
\bauthor{\bsnm{Deldjoo}, \binits{Y.}},
\bauthor{\bsnm{Noia}, \binits{T.D.}},
\bauthor{\bsnm{Merra}, \binits{F.A.}}:
\batitle{A survey on adversarial recommender systems: from attack/defense
  strategies to generative adversarial networks}.
\bjtitle{CSUR}
\bvolume{54}(\bissue{2}),
\bfpage{1}--\blpage{38}
(\byear{2021})
\end{barticle}
\endbibitem

\bibitem[\protect\citeauthoryear{Yang et~al.}{2016}]{yang2016re}
\begin{barticle}
\bauthor{\bsnm{Yang}, \binits{Z.}},
\bauthor{\bsnm{Xu}, \binits{L.}},
\bauthor{\bsnm{Cai}, \binits{Z.}},
\bauthor{\bsnm{Xu}, \binits{Z.}}:
\batitle{Re-scale adaboost for attack detection in collaborative filtering
  recommender systems}.
\bjtitle{Knowledge-Based Systems}
\bvolume{100},
\bfpage{74}--\blpage{88}
(\byear{2016})
\end{barticle}
\endbibitem

\bibitem[\protect\citeauthoryear{Ge et~al.}{2022}]{ge2022survey}
\begin{botherref}
\oauthor{\bsnm{Ge}, \binits{Y.}},
\oauthor{\bsnm{Liu}, \binits{S.}},
\oauthor{\bsnm{Fu}, \binits{Z.}},
\oauthor{\bsnm{Tan}, \binits{J.}},
\oauthor{\bsnm{Li}, \binits{Z.}},
\oauthor{\bsnm{Xu}, \binits{S.}},
\oauthor{\bsnm{Li}, \binits{Y.}},
\oauthor{\bsnm{Xian}, \binits{Y.}},
\oauthor{\bsnm{Zhang}, \binits{Y.}}:
A survey on trustworthy recommender systems.
arXiv preprint arXiv:2207.12515
(2022)
\end{botherref}
\endbibitem

\bibitem[\protect\citeauthoryear{Zhang et~al.}{2015}]{zhang2015catch}
\begin{bchapter}
\bauthor{\bsnm{Zhang}, \binits{Y.}},
\bauthor{\bsnm{Tan}, \binits{Y.}},
\bauthor{\bsnm{Zhang}, \binits{M.}},
\bauthor{\bsnm{Liu}, \binits{Y.}},
\bauthor{\bsnm{Chua}, \binits{T.-S.}},
\bauthor{\bsnm{Ma}, \binits{S.}}:
\bctitle{Catch the black sheep: unified framework for shilling attack detection
  based on fraudulent action propagation}.
In: \bbtitle{Twenty-fourth International Joint Conference on Artificial
  Intelligence}
(\byear{2015})
\end{bchapter}
\endbibitem

\bibitem[\protect\citeauthoryear{Zhang et~al.}{2018}]{zhang2018ud}
\begin{barticle}
\bauthor{\bsnm{Zhang}, \binits{F.}},
\bauthor{\bsnm{Zhang}, \binits{Z.}},
\bauthor{\bsnm{Zhang}, \binits{P.}},
\bauthor{\bsnm{Wang}, \binits{S.}}:
\batitle{Ud-hmm: An unsupervised method for shilling attack detection based on
  hidden markov model and hierarchical clustering}.
\bjtitle{Knowledge-Based Systems}
\bvolume{148},
\bfpage{146}--\blpage{166}
(\byear{2018})
\end{barticle}
\endbibitem

\bibitem[\protect\citeauthoryear{Zhang and Kulkarni}{2014}]{zhang2014detection}
\begin{bchapter}
\bauthor{\bsnm{Zhang}, \binits{Z.}},
\bauthor{\bsnm{Kulkarni}, \binits{S.R.}}:
\bctitle{Detection of shilling attacks in recommender systems via spectral
  clustering}.
In: \bbtitle{FUSION},
pp. \bfpage{1}--\blpage{8}
(\byear{2014}).
\bcomment{IEEE}
\end{bchapter}
\endbibitem

\bibitem[\protect\citeauthoryear{Cao et~al.}{2013}]{cao2013shilling}
\begin{barticle}
\bauthor{\bsnm{Cao}, \binits{J.}},
\bauthor{\bsnm{Wu}, \binits{Z.}},
\bauthor{\bsnm{Mao}, \binits{B.}},
\bauthor{\bsnm{Zhang}, \binits{Y.}}:
\batitle{Shilling attack detection utilizing semi-supervised learning method
  for collaborative recommender system}.
\bjtitle{WWW}
\bvolume{16}(\bissue{5-6}),
\bfpage{729}--\blpage{748}
(\byear{2013})
\end{barticle}
\endbibitem

\bibitem[\protect\citeauthoryear{Cheng and Hurley}{2009}]{cheng2009effective}
\begin{bchapter}
\bauthor{\bsnm{Cheng}, \binits{Z.}},
\bauthor{\bsnm{Hurley}, \binits{N.}}:
\bctitle{Effective diverse and obfuscated attacks on model-based recommender
  systems}.
In: \bbtitle{Proceedings of the Third ACM Conference on Recommender Systems},
pp. \bfpage{141}--\blpage{148}
(\byear{2009})
\end{bchapter}
\endbibitem

\bibitem[\protect\citeauthoryear{Athalye et~al.}{2018}]{athalye2018obfuscated}
\begin{bchapter}
\bauthor{\bsnm{Athalye}, \binits{A.}},
\bauthor{\bsnm{Carlini}, \binits{N.}},
\bauthor{\bsnm{Wagner}, \binits{D.}}:
\bctitle{Obfuscated gradients give a false sense of security: Circumventing
  defenses to adversarial examples}.
In: \bbtitle{ICML},
pp. \bfpage{274}--\blpage{283}
(\byear{2018}).
\bcomment{PMLR}
\end{bchapter}
\endbibitem

\bibitem[\protect\citeauthoryear{Machado et~al.}{2021}]{machado2021adversarial}
\begin{botherref}
\oauthor{\bsnm{Machado}, \binits{G.R.}},
\oauthor{\bsnm{Silva}, \binits{E.}},
\oauthor{\bsnm{Goldschmidt}, \binits{R.R.}}:
Adversarial machine learning in image classification: A survey toward the
  defender’s perspective.
CSUR
(1),
1--38
(2021)
\end{botherref}
\endbibitem

\bibitem[\protect\citeauthoryear{He et~al.}{2018}]{he2018adversarial}
\begin{bchapter}
\bauthor{\bsnm{He}, \binits{X.}},
\bauthor{\bsnm{He}, \binits{Z.}},
\bauthor{\bsnm{Du}, \binits{X.}},
\bauthor{\bsnm{Chua}, \binits{T.-S.}}:
\bctitle{Adversarial personalized ranking for recommendation}.
In: \bbtitle{SIGIR},
pp. \bfpage{355}--\blpage{364}
(\byear{2018})
\end{bchapter}
\endbibitem

\bibitem[\protect\citeauthoryear{Li et~al.}{2020}]{li2020adversarial}
\begin{bchapter}
\bauthor{\bsnm{Li}, \binits{R.}},
\bauthor{\bsnm{Wu}, \binits{X.}},
\bauthor{\bsnm{Wang}, \binits{W.}}:
\bctitle{Adversarial learning to compare: Self-attentive prospective customer
  recommendation in location based social networks}.
In: \bbtitle{WSDM},
pp. \bfpage{349}--\blpage{357}
(\byear{2020})
\end{bchapter}
\endbibitem

\bibitem[\protect\citeauthoryear{Park and Chang}{2019}]{park2019adversarial}
\begin{bchapter}
\bauthor{\bsnm{Park}, \binits{D.H.}},
\bauthor{\bsnm{Chang}, \binits{Y.}}:
\bctitle{Adversarial sampling and training for semi-supervised information
  retrieval}.
In: \bbtitle{The World Wide Web Conference},
pp. \bfpage{1443}--\blpage{1453}
(\byear{2019})
\end{bchapter}
\endbibitem

\bibitem[\protect\citeauthoryear{Tang et~al.}{2019}]{tang2019adversarial}
\begin{barticle}
\bauthor{\bsnm{Tang}, \binits{J.}},
\bauthor{\bsnm{Du}, \binits{X.}},
\bauthor{\bsnm{He}, \binits{X.}},
\bauthor{\bsnm{Yuan}, \binits{F.}},
\bauthor{\bsnm{Tian}, \binits{Q.}},
\bauthor{\bsnm{Chua}, \binits{T.-S.}}:
\batitle{Adversarial training towards robust multimedia recommender system}.
\bjtitle{IEEE Transactions on Knowledge and Data Engineering}
\bvolume{32}(\bissue{5}),
\bfpage{855}--\blpage{867}
(\byear{2019})
\end{barticle}
\endbibitem

\bibitem[\protect\citeauthoryear{Yue et~al.}{2022}]{yue2022defending}
\begin{bchapter}
\bauthor{\bsnm{Yue}, \binits{Z.}},
\bauthor{\bsnm{Zeng}, \binits{H.}},
\bauthor{\bsnm{Kou}, \binits{Z.}},
\bauthor{\bsnm{Shang}, \binits{L.}},
\bauthor{\bsnm{Wang}, \binits{D.}}:
\bctitle{Defending substitution-based profile pollution attacks on sequential
  recommenders}.
In: \bbtitle{Proceedings of the 16th ACM Conference on Recommender Systems},
pp. \bfpage{59}--\blpage{70}
(\byear{2022})
\end{bchapter}
\endbibitem

\bibitem[\protect\citeauthoryear{Hidano and
  Kiyomoto}{2020}]{hidano2020recommender}
\begin{bchapter}
\bauthor{\bsnm{Hidano}, \binits{S.}},
\bauthor{\bsnm{Kiyomoto}, \binits{S.}}:
\bctitle{Recommender systems robust to data poisoning using trim learning.}
In: \bbtitle{ICISSP},
pp. \bfpage{721}--\blpage{724}
(\byear{2020})
\end{bchapter}
\endbibitem

\bibitem[\protect\citeauthoryear{Zhang et~al.}{2017}]{zhang2017robust}
\begin{barticle}
\bauthor{\bsnm{Zhang}, \binits{F.}},
\bauthor{\bsnm{Lu}, \binits{Y.}},
\bauthor{\bsnm{Chen}, \binits{J.}},
\bauthor{\bsnm{Liu}, \binits{S.}},
\bauthor{\bsnm{Ling}, \binits{Z.}}:
\batitle{Robust collaborative filtering based on non-negative matrix
  factorization and r1-norm}.
\bjtitle{Knowledge-based systems}
\bvolume{118},
\bfpage{177}--\blpage{190}
(\byear{2017})
\end{barticle}
\endbibitem

\bibitem[\protect\citeauthoryear{Yu et~al.}{2017}]{yu2017novel}
\begin{barticle}
\bauthor{\bsnm{Yu}, \binits{H.}},
\bauthor{\bsnm{Gao}, \binits{R.}},
\bauthor{\bsnm{Wang}, \binits{K.}},
\bauthor{\bsnm{Zhang}, \binits{F.}}:
\batitle{A novel robust recommendation method based on kernel matrix
  factorization}.
\bjtitle{Journal of Intelligent \& Fuzzy Systems}
\bvolume{32}(\bissue{3}),
\bfpage{2101}--\blpage{2109}
(\byear{2017})
\end{barticle}
\endbibitem

\bibitem[\protect\citeauthoryear{Smith and Linden}{2017}]{smith2017two}
\begin{barticle}
\bauthor{\bsnm{Smith}, \binits{B.}},
\bauthor{\bsnm{Linden}, \binits{G.}}:
\batitle{Two decades of recommender systems at amazon. com}.
\bjtitle{Ieee internet computing}
\bvolume{21}(\bissue{3}),
\bfpage{12}--\blpage{18}
(\byear{2017})
\end{barticle}
\endbibitem

\bibitem[\protect\citeauthoryear{Gomez-Uribe and Hunt}{2015}]{gomez2015netflix}
\begin{barticle}
\bauthor{\bsnm{Gomez-Uribe}, \binits{C.A.}},
\bauthor{\bsnm{Hunt}, \binits{N.}}:
\batitle{The netflix recommender system: Algorithms, business value, and
  innovation}.
\bjtitle{ACM Transactions on Management Information Systems (TMIS)}
\bvolume{6}(\bissue{4}),
\bfpage{1}--\blpage{19}
(\byear{2015})
\end{barticle}
\endbibitem

\bibitem[\protect\citeauthoryear{WU et~al.}{2014}]{wu2014survey}
\begin{barticle}
\bauthor{\bsnm{WU}, \binits{Z.}},
\bauthor{\bsnm{WANG}, \binits{Y.}},
\bauthor{\bsnm{CAO}, \binits{J.}}:
\batitle{A survey on shilling attack models and detection techniques for
  recommender systems}.
\bjtitle{Chinese Sci Bull}
\bvolume{59}(\bissue{7}),
\bfpage{551}--\blpage{560}
(\byear{2014})
\end{barticle}
\endbibitem

\bibitem[\protect\citeauthoryear{Zhang et~al.}{2020}]{zhang2020attacks}
\begin{bchapter}
\bauthor{\bsnm{Zhang}, \binits{J.}},
\bauthor{\bsnm{Xu}, \binits{X.}},
\bauthor{\bsnm{Han}, \binits{B.}},
\bauthor{\bsnm{Niu}, \binits{G.}},
\bauthor{\bsnm{Cui}, \binits{L.}},
\bauthor{\bsnm{Sugiyama}, \binits{M.}},
\bauthor{\bsnm{Kankanhalli}, \binits{M.}}:
\bctitle{Attacks which do not kill training make adversarial learning
  stronger}.
In: \bbtitle{ICML},
pp. \bfpage{11278}--\blpage{11287}
(\byear{2020}).
\bcomment{PMLR}
\end{bchapter}
\endbibitem

\bibitem[\protect\citeauthoryear{Yuan et~al.}{2019}]{yuan2019adversarial}
\begin{bchapter}
\bauthor{\bsnm{Yuan}, \binits{F.}},
\bauthor{\bsnm{Yao}, \binits{L.}},
\bauthor{\bsnm{Benatallah}, \binits{B.}}:
\bctitle{Adversarial collaborative neural network for robust recommendation}.
In: \bbtitle{SIGIR},
pp. \bfpage{1065}--\blpage{1068}
(\byear{2019})
\end{bchapter}
\endbibitem

\bibitem[\protect\citeauthoryear{Raghunathan
  et~al.}{2019}]{raghunathan2019adversarial}
\begin{botherref}
\oauthor{\bsnm{Raghunathan}, \binits{A.}},
\oauthor{\bsnm{Xie}, \binits{S.M.}},
\oauthor{\bsnm{Yang}, \binits{F.}},
\oauthor{\bsnm{Duchi}, \binits{J.C.}},
\oauthor{\bsnm{Liang}, \binits{P.}}:
Adversarial training can hurt generalization.
arXiv preprint arXiv:1906.06032
(2019)
\end{botherref}
\endbibitem

\end{thebibliography}

\end{document}